\documentclass{article}

\usepackage[preprint]{neurips_2026}

\usepackage[utf8]{inputenc} 
\usepackage[T1]{fontenc}    
\usepackage{hyperref}       
\usepackage{url}            
\usepackage{booktabs}       
\usepackage{amsfonts}       
\usepackage{nicefrac}       
\usepackage{microtype}      
\usepackage{xcolor}         

\usepackage{graphicx}
\usepackage{amsmath}
\usepackage{algorithm}
\usepackage{algpseudocode}
\usepackage{multirow}
\usepackage{comment}
\usepackage{subcaption}
\usepackage{pifont}
\usepackage[super]{nth}
\usepackage[table]{xcolor}

\definecolor{ForestGreen}{RGB}{34,139,34}

\title{EDMA: Entropy-Driven Multimodal Answering}

\author{%
  Emanuele Mezzi \\
  Vrije Universiteit Amsterdam \\
  \texttt{e.mezzi@vu.nl}
  \And
  Gertjan Burghouts \\ 
  TNO \\
  \texttt{gertjan.burghouts@tno.nl}
  \And
  Fabio Massacci \\
  Vrije Universiteit Amsterdam \\
  University of Trento \\ 
  \texttt{fabio.massacci@ieee.org} \\
  \And
  Mengyuan Zhang \\
  Vrije Universiteit Amsterdam \\
  \texttt{m.zhang@vu.nl}
}

\begin{document}

\maketitle

\begin{abstract}
Multimodal question answering (MMQA) is based on integrating heterogeneous data sources, selectively leveraging relevant modalities, and ignoring distractors. Existing approaches based on Multimodal Large Language Models (MLLMs) improve performance by splitting the task into intermediate steps, but do not quantify how each modality contributes to reaching the final answer. We argue that quantifying the effect that each modality has on the answering process improves process traceability and final performance. Thus, we propose Entropy-Driven Multimodal Answering (EDMA), which uses logical entropy to formalise multimodal QA as an iterative process of entropy reduction and to quantify the contribution of each modality in distinguishing correct from incorrect candidate answers. This enables the construction of \emph{quantifiable answering trajectories}, where each step is associated with a measurable entropy reduction. EDMA introduces (i) modality-conditioned partitioning to estimate unimodal entropy reduction, (ii) entropy-driven multimodal fusion to capture complementary information across modalities, and (iii) entropy-based selection of the answering trajectory that minimises the most logical entropy. Evaluated across standard MMQA benchmarks, EDMA outperforms competitive prompting baselines, with average gains of 12.8 F1 and 4.6 LLM-J points over Planning Prompting and 18.6 F1 and 8.3 LLM-J points over Chain-of-Thought. Moreover, EDMA outperforms self-consistency baselines and is competitive with the multi-agent system baseline. Moreover, we show that higher entropy correlates with more false positives and, through controlled trajectory selection, demonstrate that logical entropy serves as an effective, monotonic selector for correct reasoning paths, validating entropy level as a reliable proxy for performance. \footnote{Code will be made publicly available with the final camera-ready version.}
\end{abstract}

\begin{figure}[h]
\centering
\includegraphics[width=\linewidth]{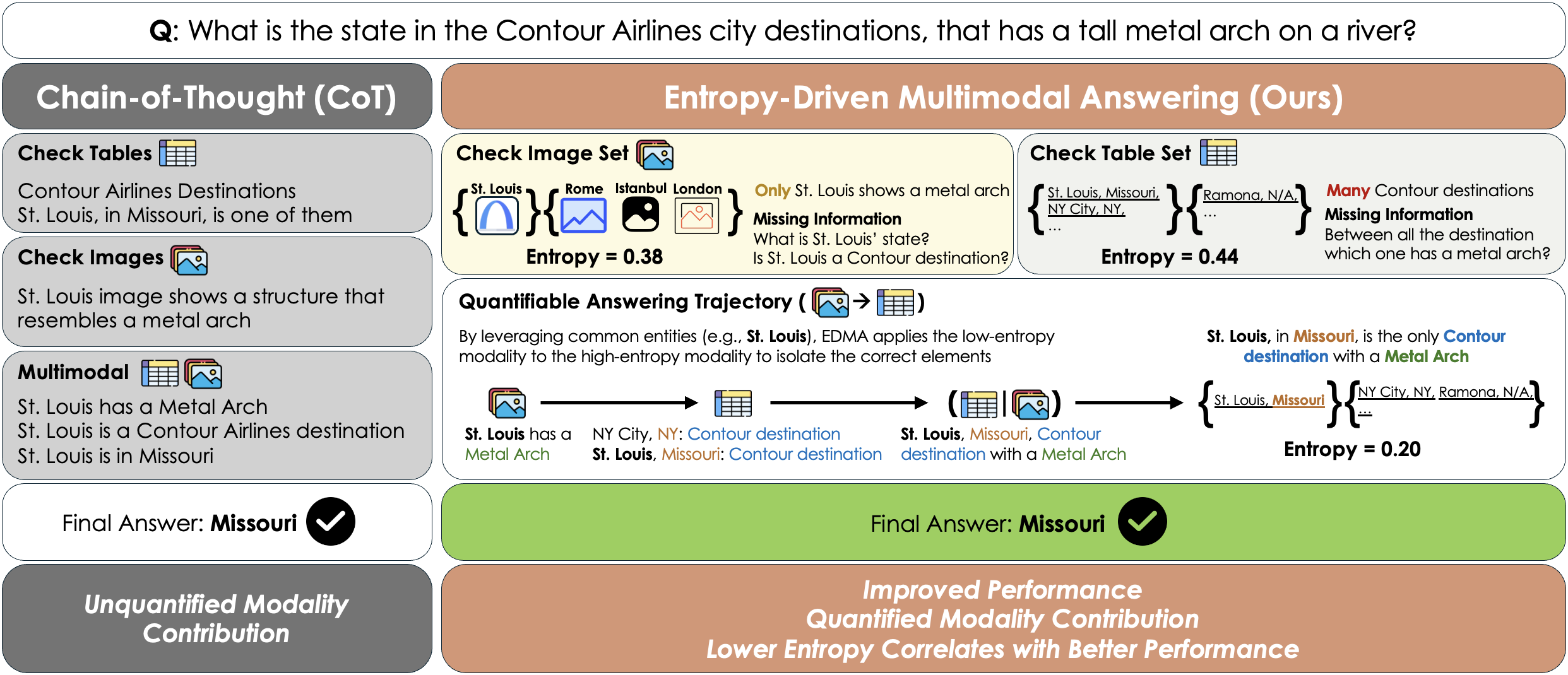}
\caption{Comparison between CoT and EDMA. While they both structure QA in intermediate steps, EDMA generates quantifiable answering trajectories, allowing for improved performance and quantified modality contribution.}
\label{fig:us_and_them}
\end{figure}

\section{Introduction}
Multimodal question answering (MMQA) requires the integration of heterogeneous data sources such as text, images, and tables \cite{talmor2021multimodalqa, luo2023unifying, foroutan2025wikimixqa}. For example, answering the question \emph{``What is the state in the Contour Airlines city destinations, that has a tall metal arch on a river?''} requires identifying visual evidence of a metal arch in images while simultaneously retrieving structured information from tables to determine the destinations of the airline, and combining the evidence. 

Recent Multimodal Large Language Models (MLLMs) have shown strong performance on such tasks by leveraging unified representations across modalities \cite{gao2023g, zhang2025towards}, and are currently being employed to solve domain-specific problems \cite{li2025chemvlm, yang2025lidar, xia2024sportu, wang2024t, zhao2025r1}. Several techniques have been proposed to improve multimodal inference, most notably Chain-of-Thought (CoT) prompting \cite{wei2022chain, zhang2023multimodal, zheng2023ddcot}, which decomposes complex tasks into intermediate steps. Building on this paradigm, extensions such as compositional prompting \cite{mitra2024compositional}, knowledge-augmented reasoning \cite{mondal2024kam}, PEAR-CoT \cite{li2025multimodal}, frameworks such as Chain-of-Action (CoA) \cite{pan2024chain} and Visualization-of-Thought (VoT) \cite{wu2024mind} have been proposed to address limitations in compositionality, grounding, and spatial awareness. To further improve the inference process, recent work has incorporated uncertainty quantification, enabling models to adaptively seek information or refine intermediate steps \cite{hu2024uncertainty, yin2024reasoning, liang2024introspective}. However, while these approaches improve robustness by structuring the inference process in intermediate steps, augmenting contextual information, and guaranteeing higher robustness, they do not provide explicit stepwise quantification of how individual modalities and their fusion contribute to reducing uncertainty over the space of candidate answers. As a result, it remains unclear how different modalities contribute to the inference process, limiting interpretability and controllability in multimodal settings.

To address these limitations, we propose Entropy-Driven Multimodal Answering (EDMA), a methodology that models QA as a process of iterative entropy reduction over candidate answers. As Figure \ref{fig:us_and_them} shows, EDMA quantifies the impact of incorporating a specific modality and fusing modalities by measuring how this refines the space of possible answers. Compared to CoT, which splits the answering process in steps without quantifying the contribution of each modality, EDMA builds \emph{quantifiable answering trajectories}, where each step and each modality is associated with a measurable contribution in the reduction of entropy. At the core of EDMA lies logical entropy \cite{ellerman2021introduction}, which measures distinctions between possible outcomes. This formulation is particularly well-suited for multimodal QA, where the objective is to progressively separate correct from incorrect candidate answers based on evidence from different modalities. By framing answering as partition refinement, EDMA treats each piece of evidence as a condition that splits the current answer space into two subsets of answers: those that satisfy the evidence and those that do not. This systematic partitioning provides a natural and interpretable mechanism for tracking exactly how information from each modality contributes to narrowing down the final prediction.

We evaluate EDMA on challenging multimodal QA benchmarks \cite{talmor2021multimodalqa, hannan2020manymodalqa} and show that it outperforms best-practice prompting baselines and, on average, state-of-the-art multi-agent systems, while providing traceable uncertainty estimates that guide modality selection and the answering process. Our contributions are as follows:

\begin{itemize}
    \item EDMA, a methodology that formalizes multimodal QA as modality-conditioned entropy minimization over the space of alternatives.
    
    \item A modality-conditioned entropy reduction objective that enables explicit measurement of each modality’s contribution.
    
    \item We establish the relation between entropy minimization and lower false positives (FPs) levels, thus validating entropy minimization as a proxy for performance. 
\end{itemize}

\def\answerer{\ensuremath{LLM}}
\def\question{\ensuremath{Q}}                          
\def\modalities{\ensuremath{M}}                        
\def\modality{\ensuremath{m}}                           
\def\answer{\ensuremath{a^*}}
\def\startmodalities{\ensuremath{M_s}}                 
\def\analysemodalites{\ensuremath{M_{a}}}
\def\targetmodality{\ensuremath{m_t}}                  
\def\conditioningmodalities{\ensuremath{M_c}}          
\def\prompt{\ensuremath{Prompt}}

\def\elements{\ensuremath{\Omega_m}}                   
\def\element{\ensuremath{\omega}}                      

\def\requirements{\ensuremath{R}}                      
\def\requirement{\ensuremath{r}}  
\def\negrequirement{\ensuremath{\neg \requirement}}

\def\partialanswers{\ensuremath{E_{\modality}}}     
\def\negpartialanswers{\ensuremath{\neg \partialanswers}}
\def\partialanswer{\ensuremath{e}}

\def\unipartition{\ensuremath{P_{\modality}^{\requirement}}}               
\def\targetpartition{\ensuremath{P_{\targetmodality|\modality}^{\requirement,\partialanswers}}}    

\def\unientropy{\ensuremath{H(P_{\modality}^{\requirement})}}              
\def\targetentropy{\ensuremath{H(\targetpartition)}}

\def\targetentropygeneric{\ensuremath{H(P_{\modality}^{\requirement,\partialanswers})}}  
\def\deltaentropy{\ensuremath{\Delta H}}    

\def\omegafilling{\ensuremath{\Omega_{\modality}^{\requirement}}}          
\def\omeganotfilling{\ensuremath{\Omega_{\modality}^{\negrequirement}}}  
\def\omegafillingtarget{\ensuremath{\Omega_{\targetmodality}^{\requirement}}}
\def\omegatarget{\ensuremath{\Omega_{\targetmodality}^{\requirement, \partialanswers}}}       
\def\omegatargetneg{\ensuremath{\Omega_{\targetmodality}^{\requirement,\negpartialanswers}}} 

\def\imagemod{\ensuremath{I}}
\def\tablemod{\ensuremath{T}}

\def\statecriteria{\ensuremath{S}}
\def\metalarchcriteria{\ensuremath{A}}
\def\contourdestinationcriteria{\ensuremath{C}}

\def\trajectoryset{\ensuremath{\mathcal{T}}}
\def\optimaltrajectory{\ensuremath{t^*}}
\def\finalpartialanswers{\ensuremath{E^*}}
\def\finalelements{\ensuremath{\Omega^*}}

\section{Entropy-Driven Multimodal Answering (EDMA)}
\label{sec:methodology}

\subsection{Problem formulation}
The problem setting involves one question $\question$, the set of modalities available to answer the question denoted by $\modalities$ and, for each $\modality \in \modalities$, the elements associated with that modality denoted by $\elements$, and one Answerer denoted by $\answerer$. To answer each question, the $\answerer$ must find the modalities of interest, find the elements within those modalities that contain the necessary information to answer the question, and then combine the information to provide a cross-modal answer, denoted by $\answer$. 

\subsection{Overview of EDMA}

\begin{figure}[h]
    \centering
    \includegraphics[width=\linewidth]{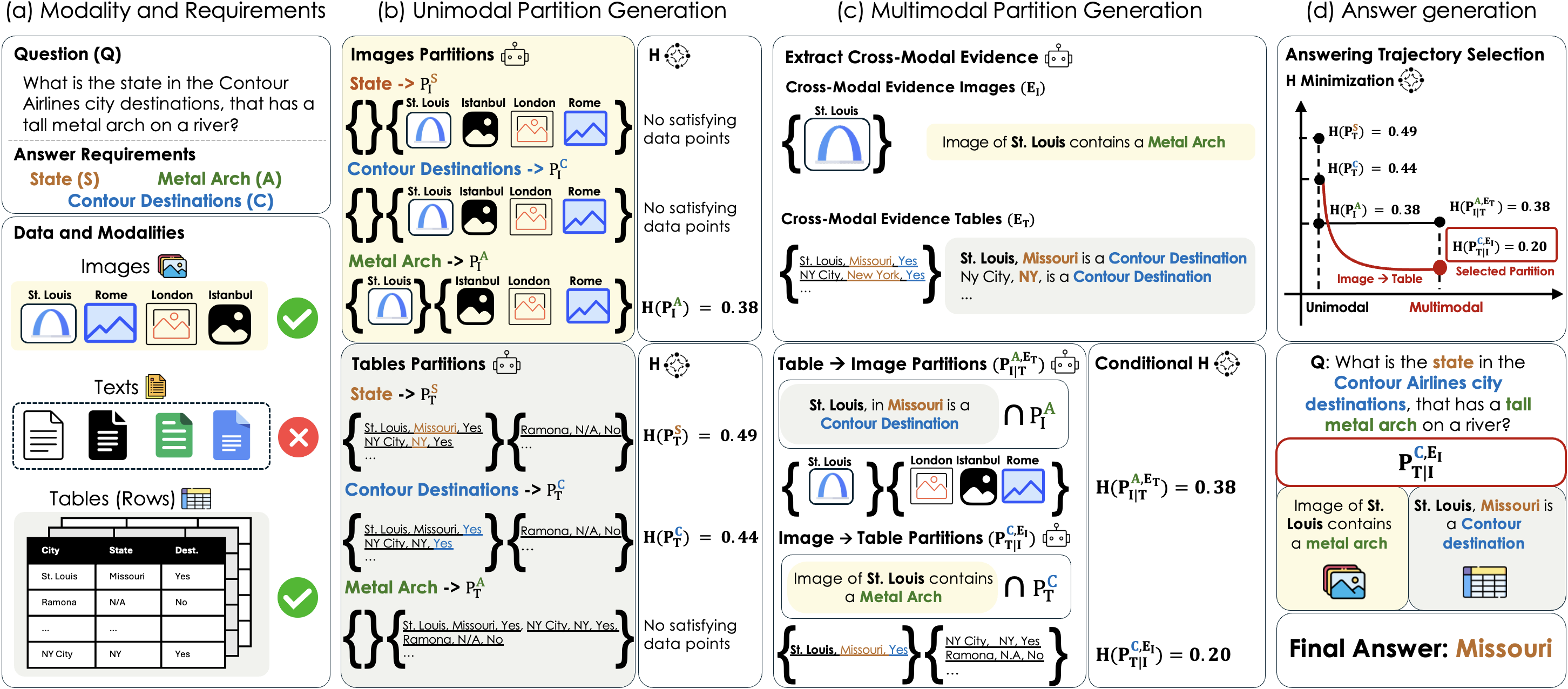}
    \caption{EDMA is composed of four stages: (a) modality prediction and extraction of answer requirements, (b) generation of unimodal partitions and calculation of unimodal logical entropy, (c) generation of multimodal partitions and calculation of multimodal logical entropy, (d) selection of the answering trajectory which minimises the logical entropy and generation of the answer.}
    \label{fig:methodology}
\end{figure}

Figure \ref{fig:methodology} shows that EDMA is composed of four stages: (a) prediction of the modalities to be used and extraction of answer requirements from the question, (b) generation of unimodal partitions and calculation of unimodal logical entropy, (c) generation of the multimodal partitions with paired multimodal logical entropy, and (d) answer generation by selecting the answering trajectory that minimized the most the logical entropy. The process allows to reduce at each step the number of elements which represent the possible alternatives by checking which ones fit the requirements extracted from the question, returning an answer based on the answering trajectory that minimised the most the logical entropy. In Algorithm \ref{alg:edma} in Appendix \ref{appendix:sec:algorithm}, we describe EDMA's answering process.

\subsection{Extraction of Answer Requirements}
Supposing we want to answer the question \emph{``What is the state in the Contour Airlines city destinations, that has a tall metal arch on a river?''}, EDMA begins by extracting the set of answer requirements, representing the characteristics that the final answer $\answer$ must satisfy. We denote the answer requirements as $\requirements$. As shown in Figure \ref{fig:methodology}, the $\answerer$ generates $\requirements$ by identifying the key elements of the question. In this specific case, the final answer has to: (i) be a \emph{State}, (ii) be a \emph{destination of Contour Airlines}, and (iii) be characterized by a \emph{Metal Arch}. We denote the extraction of $\requirements$ as $\requirements = LLM(\prompt_{R}, \question)$, where the $\answerer$ receives in input the user instructions $\prompt_{R}$ and the question $\question$. In this case, $\requirements = \{S, C, A\}$, where we use $S$ to point to State, $C$ for Contour Destination, and $A$ for Metal Arch. 

\subsection{Prediction of Starting Modalities}
EDMA predicts the subset of modalities $\startmodalities \subseteq \modalities$ representing the modalities with which to start the analysis. This phase does not exclude the possibility of incorporating additional modalities later in the answering process. The $\answerer$ receives the input, the user instructions $\prompt_{\startmodalities}$, the question $\question$, and for each modality $\modality$, the corresponding set of elements $\elements$. It is then tasked with predicting the combination of modalities that is most likely to contain the elements needed to reach the correct answer. This operation can be summarized as $\startmodalities = LLM(\prompt_{\startmodalities}, \question, \{\Omega_m,  \modality \in \modalities\})$. The modalities $\startmodalities$ are then added to $\analysemodalites$, which is the list that keeps track of the modalities used for the analysis. In Figure \ref{fig:methodology}, the predicted modalities are \emph{Table (T)} and \emph{Image (I)}. 

\subsection{Entropy-Driven Answer Search}
To identify the correct answer in a multimodal setting, EDMA models inference as an entropy-driven search over the space of candidate answers. It quantifies how individual modalities and their interactions reduce uncertainty over the answer space. To this end, EDMA first induces partitions over the answer space using unimodal evidence and then refines these partitions through sequential cross-modal conditioning. This process enables systematic measurement of entropy reduction from both unimodal and multimodal evidence.

\paragraph{Quantifying Uncertainty of Answer Partitions}
We adopt logical entropy \cite{ellerman2021introduction}, which measures uncertainty in terms of distinctions induced by a partition of a candidate set. Let $S$ denote a finite set of candidate elements and let $P_S = \{B_1, \dots, B_n\}$ be a partition of $S$ into disjoint non-empty blocks. To ensure compatibility across open-source and closed-source models without requiring calibrated probability estimates from the underlying model, we assume a uniform distribution over $S$, such that each block has probability mass $p_i = \frac{|B_i|}{|S|}$. Logical entropy is defined as $H(P_S) = 1 - \sum_{i=1}^{n} p_i^2$, and can be interpreted as the probability that two independently sampled elements from $S$ belong to different blocks.

For example, consider $|S| = 10$ and a binary partition $P_S = \{S_{\requirement}, S_{\neg \requirement}\}$ induced by a requirement $\requirement$. The set $S_{\requirement}$ contains elements whose information overlaps with $\requirement$, while $S_{\neg \requirement}$ contains elements that do not. If $|S_{\requirement}| = 5$ and $|S_{\neg \requirement}| = 5$, then $H(P_S) = 0.50$, corresponding to maximal entropy for a binary partition under a uniform prior. If instead $|S_{\requirement}| = 1$ and $|S_{\neg \requirement}| = 9$, then $H(P_S) = 1 - (0.1^2 + 0.9^2) = 0.18$, reflecting reduced uncertainty due to the increased concentration of mass.

In our multimodal QA setting, the answering process is modeled as a sequence of partition refinements over $S$, where information extracted from the elements of each modality introduces distinctions that separate relevant from irrelevant candidates. We restrict attention to valid modality-induced refinements where the predicate-satisfying subset is non-empty, $S_{\requirement} \neq \emptyset$, ensuring that entropy reduction is grounded in non-vacuous evidence.

Logical entropy thus quantifies the effectiveness of each modality in refining the answer space, allowing to prioritize modalities that maximize entropy reduction, guiding the system toward increasingly discriminative evidence.

\paragraph{Generation of Unimodal Partitions} 
EDMA analyses the elements $\elements$ of each modality $\modality \in \analysemodalites$. For each element $\element \in \elements$, the $\answerer$ evaluates, for each answer requirement $\requirement \in \requirements$, whether $\element$ contains information overlapping with $\requirement$. This operation can be formally expressed as $LLM(\prompt_{Check}, \omega, \requirement)$. Based on this evaluation, EDMA constructs, for each set $\elements$ and requirement $\requirement$, a partition $\unipartition = \{\omegafilling, \omeganotfilling\}$, where $\omegafilling$ contains elements whose information overlaps with $\requirement$, and $\omeganotfilling$ contains elements that do not. Under the assumption of uniform distribution over elements in $\elements$, the logical entropy of $\unipartition$ is: 

\begin{equation}
\unientropy = 1 - \left( \left(\frac{|\omegafilling|}{|\elements|}\right)^2 + \left(\frac{|\omeganotfilling|}{|\elements|}\right)^2 \right)
\end{equation}

\def\omegaimagemod{\ensuremath{\Omega_{\imagemod}}}
\def\omegatablemod{\ensuremath{\Omega_{\tablemod}}}

\def\unipartitionimagearch{\ensuremath{P_{\imagemod}^{\text{\metalarchcriteria}}}}
\def\entropunipartitionimagearch{\ensuremath{H(\unipartitionimagearch)}}

\def\unipartitiontablestate{\ensuremath{P_{\tablemod}^\text{\statecriteria}}}
\def\entropyunipartitiontablestate{\ensuremath{H(\unipartitiontablestate)}}

\def\unipartitiontablecontour{\ensuremath{P_{\tablemod}^\text{\contourdestinationcriteria}}}
\def\entropyunipartitiontablecontour{\ensuremath{H(\unipartitiontablecontour)}}

This formulation expresses logical entropy as a function of partition cardinalities, enabling estimation without relying on confidence estimates from the $\answerer$. Logical entropy measures the distinction between elements that satisfy a given condition and those that do not. It attains its global minimum of $0$ when either $\omegafilling = \emptyset$ or $\omeganotfilling = \emptyset$. Otherwise, it is strictly positive and is minimized when one subset is highly concentrated (i.e., when a single candidate is clearly distinguished).

In Figure \ref{fig:methodology}, unimodal partitions illustrate how different modalities contribute to distinguishing relevant from irrelevant candidates. The image-based partition $\unipartitionimagearch$ yields a relatively low entropy ($0.38$), indicating that the visual modality already provides a strong separation by isolating the candidate satisfying the requirement $\metalarchcriteria$. In contrast, the table-based partitions exhibit higher entropy values (e.g., $0.49$ for the $S$ requirement), reflecting a weaker discriminative effect when considered in isolation. Since lower entropy corresponds to sharper distinctions among candidates, this comparison indicates that the image modality is more informative at this stage for narrowing the answer space. EDMA leverages this signal to prioritize modalities that induce stronger reductions in uncertainty, guiding the answering process toward partitions that more effectively constrain the set of plausible answers.

\paragraph{Generation of Multimodal Partitions}
Once unimodal partitions $\unipartitionimagearch$, $\unipartitiontablestate$, $\unipartitiontablecontour$ are constructed, EDMA proceeds to generate multimodal partitions. This step leverages information extracted from multiple modalities to refine candidate elements, progressively reducing logical entropy, with the aim of isolating the correct candidate(s), filtering out the alternatives that are inconsistent with cross-modal evidence, which is information extracted from one modality used to reduce the answer space in another modality. In a multimodal setting, the final answer typically resides in a specific modality, referred to as the \emph{target modality}, denoted by $\targetmodality$. However, the target modality is not known in advance. To identify it, EDMA systematically treats each modality in turn as a candidate target modality $\targetmodality$, while the remaining modalities $\conditioningmodalities = \analysemodalites \setminus \{\targetmodality\}$ serve as conditioning modalities. 

For each candidate $\targetmodality$, EDMA applies information from all conditioning modalities sequentially. For each conditioning modality $\modality \in \conditioningmodalities$ and the subset $\Omega_{\modality}^{\requirement'}$ containing elements that satisfy a requirement $\requirement' \in \requirements$, EDMA extracts the cross-modal evidence set:

\begin{equation}
\partialanswers = \Big\{\, \partialanswer \;\big|\; \partialanswer = LLM(\prompt_{Extract}, \question, \element),\ \element \in \Omega_{\modality}^{\requirement'} \,\Big\}
\end{equation}

Each piece of evidence represents information from the conditioning modality $\modality$ and has the same function as the answer requirements. EDMA applies the elements of this set sequentially to iteratively refine the candidates in the target modality $\targetmodality$ that satisfy requirements $\requirement \in \requirements$, where $\requirement \neq \requirement'$. This allows evidence from one modality (e.g., visual) that satisfies one requirement to be used to filter candidates in another modality (e.g., tabular) that satisfy a distinct requirement.

Specifically, for a given requirement $\requirement \in \requirements$, the target subset $\omegafillingtarget$ is updated sequentially for each conditioning modality. At each step, EDMA generates a partition $\targetpartition = \{\omegatarget, \omegatargetneg\}$ of the elements of modality $\targetmodality$ consistent with requirement $\requirement$, where $\omegatarget$ contains the elements that are also consistent with the evidence $\partialanswers$ from the current conditioning modality, and $\omegatargetneg$ contains those that are not. Only the subset $\omegatarget$ is carried forward to the next iteration, while $\omegatargetneg$ is excluded from further conditioning, as it cannot lead to the correct answer. This sequential conditioning progressively narrows the answer space of $\targetmodality$, increasing the distinction among the remaining candidates, thus decreasing logical entropy.

By evaluating the multimodal logical entropy after each step, EDMA identifies which $\targetmodality$ ultimately yields the partition with the lowest entropy, where distinctions among candidates are maximized, and uncertainty is minimized. Conditional or multimodal logical entropy is computed analogously to the unimodal case:

\begin{equation}
H\Big(\targetpartition\Big) = 1 - \left( 
\left(\frac{|\omegatarget|}{|\Omega_{\targetmodality}|}\right)^2 +
\left(\frac{|\omegatargetneg|}{|\Omega_{\targetmodality}|}\right)^2
\right)
\end{equation}

\def\partialanswersimages{\ensuremath{E_{\text{\imagemod}}}}
\def\partialanswerstables{\ensuremath{E_{\text{\tablemod}}}}

\def\multipartitionstableimagearch{\ensuremath{P_{\text{\imagemod|\tablemod}}^{\text{\metalarchcriteria}, \partialanswerstables}}}
\def\entropymultipartitionstableimagearch{\ensuremath{H(\multipartitionstableimagearch)}}

\def\multipartitionsimagetablecontour{\ensuremath{P_{\text{\tablemod|\imagemod}}^{\text{\contourdestinationcriteria}, \partialanswersimages}}}

\def\entropymultipartitionsimagetablecontour{\ensuremath{H(\multipartitionsimagetablecontour)}} 

In the example in Figure \ref{fig:methodology}, cross-modal conditioning yields asymmetric effects depending on the direction of information flow. When table-derived constraints ($\partialanswerstables$) are applied to the image modality ($\unipartitionimagearch$), entropy remains unchanged ($0.38$), indicating limited additional discriminative power beyond visual cues. In contrast, conditioning the table modality ($\unipartitiontablecontour$) using visual evidence ($\partialanswersimages$) significantly reduces entropy ($0.44 \rightarrow 0.20$), showing that the image provides high-value filtering information that effectively disambiguates table entries, and showing that EDMA does not treat modalities symmetrically, but instead identifies directional entropy reduction, where certain modalities act as stronger selectors of relevant candidates if the derived information is applied in the right order. 

\subsection{Answer Generation}

\paragraph{Trajectory Selection via Entropy Minimization}
The systematic generation of unimodal and multimodal partitions effectively maps a set of \emph{quantifiable answering trajectories} $(\trajectoryset)$, representing paths in the search process. Each trajectory provides a quantifiable history of how information is fused and how logical entropy changed after the modality fusion. This allows for a modality-level attribution of the logical entropy variation. A trajectory may remain within a unimodal setting ($\modality \to \modality$) or bridge modalities to form a multimodal trajectory ($\modality \to \targetmodality$), where the first modality conditions the second $(\targetmodality|\modality)$. We denote $\deltaentropy = \frac{H(P_{\targetmodality}^{\requirement}) - \targetentropy}{H(P_{\targetmodality}^{\requirement})}$, the relative change between unimodal entropy and multimodal entropy for modality $\targetmodality$. If $\deltaentropy > 0$, modality fusion increases the distinguishability among candidates, thereby reducing uncertainty. If $\deltaentropy \leq 0$, the conditional modality provides redundant or non-discriminative information for that specific answering path.

In Figure \ref{fig:methodology}, EDMA identifies four answering trajectories, of which two are unimodal and two are multimodal. The unimodal trajectories are represented by $\unipartitionimagearch$ (initial $H = 0.38$) and $\unipartitiontablecontour$ (initial $H = 0.44$). The Table-Image Trajectory starts with partition $\unipartitionimagearch$ and ends with partition $\multipartitionstableimagearch$, where table data regarding destinations conditions the image set, resulting in a final entropy of $H = 0.38$ with $\deltaentropy = 0\%$. In contrast, the Image-Table Trajectory, where visual evidence of the \emph{Metal Arch} conditions the table rows, starts with partition $\unipartitiontablecontour$ and ends with partition $\multipartitionsimagetablecontour$, reducing logical entropy from $H = 0.44$ to $H = 0.20$, achieving a significant entropy decrease of $\deltaentropy = 54\%$, a reduction attributed to the image modality's influence on the table modality. 

As EDMA prioritizes the trajectory whose final partition yields the lowest logical entropy $(\optimaltrajectory)$, it selects the Image–Table Trajectory, isolating the table entry corresponding to the city satisfying requirement $C$, which is also associated with an image from which EDMA extracts visual evidence of the arch. The multimodal partition at the endpoint of $\optimaltrajectory$ is chosen as the basis for the final answer.

\paragraph{Final Inference}
Once the optimal trajectory, modality, and partition are selected, the $\answerer$ receives the question $\question$, the satisfied requirements, the cross-modal evidence $\finalpartialanswers$ (e.g., the context of the arch), and the relevant elements $\finalelements$ that satisfy both the requirements and the cross-modal evidence (e.g., the specific row for St. Louis). Based on this information, the system returns the final answer $\answer$ (e.g., \emph{Missouri}). If the $\answerer$ cannot produce an answer (i.e., $\answer = \text{None}$), EDMA selects a new set of modalities $\startmodalities \subseteq \modalities \setminus \analysemodalites$, where the newly selected modalities are not contained in the already analysed modality set. EDMA then repeats the process: generating new partitions, recomputing entropy deltas, and attempting to produce the answer with the newly acquired information. The iteration continues until a valid answer is obtained or all modalities are exhausted.

Crucially, by constructing and selecting among quantifiable answering trajectories, EDMA provides a principled mechanism to guide cross-modal inference. Even when SOTA approaches arrive at the same final answer, EDMA makes explicit how that answer is obtained, enabling quantifiable modality-level attribution.

\section{Experiments}
\label{sec:experiments}

\subsection{Experimental Setup}
We evaluate our methodology across a diverse set of SOTA LLMs, including both closed source models, such as GPT-5.2 \cite{openai2025gpt52systemcard}, Kimi-K2.5 \cite{team2026kimi}, Nova-2-Lite-v1 (Nova-2-Lite) \cite{aws2025nova2lite}, and Claude-Sonnet-4.6 (Claude-4.6) \cite{anthropic2026sonnet46}, and open source models, such as Qwen3-VL-235B-A22B (Qwen3) \cite{bai2025qwen3}, and Nemotron-Nano-12B-v2 (Nemotron) \cite{nvidia2025nemotron12bvl}. Appendix \ref{appendix:sec:access_computational_resources} reports how to access models and the resources needed to run them, while Appendix \ref{appendix:sec:prompts} reports the prompts to implement EDMA's operations.

\paragraph{Baselines} 
We compare our approach against SOTA multimodal QA system MAMMQA \cite{anvekar2025rethinking} and against several common best-practice prompting strategies: Direct Prompting (DP), where the model generates an answer without explicit intermediate steps, Chain-of-Thought (CoT) \cite{wei2022chain}, which encourages step-by-step generation of intermediate steps within a single generation, and Planning Prompting (PP) \cite{wang2023plan}, where the model first generates a plan and subsequently executes it, CoT-SC \cite{wang2022self} and PP-SC which improve the greedy versions of CoT and PP by generating multiple reasoning chains and returning the majority-vote final answer. 

\paragraph{Datasets} 
We evaluate our methodology on MultiModalQA \cite{talmor2021multimodalqa} and ManyModalQA \cite{hannan2020manymodalqa} datasets which remain, to the best of our knowledge, among the most complex benchmarks for multimodal QA, requiring cross-modal integration across text, tables, and images. They both require deep joint inference over diverse data formats to arrive at a final answer, and thus provide a rigorous testbed for evaluating a model's ability to integrate and reason over heterogeneous information sources.

\paragraph{Evaluation Metrics} 
We evaluate all methods using Exact Match (EM) and F1 score, following standard practice in multimodal QA \cite{talmor2021multimodalqa, yu2023unified, lim2024unirag}. EM is a strict binary metric that assigns a score of 1 if the predicted answer exactly matches the ground-truth string and 0 otherwise. The F1 score measures token-level overlap between the predicted and reference answers, balancing precision and recall. Because EM and F1 can be overly penalizing for semantically correct open-ended generations, in our evaluation we also include an LLM-Judge (LLM-J) metric \cite{majumdar2024openeqa}.

\subsection{EDMA Performance Evaluation}
Here we report the performance of EDMA across benchmarks. We first compare EDMA to standard prompting baselines (DP, CoT, and PP) and then we evaluate EDMA against budget-matched baselines.

\subsubsection{Comparison between EDMA and Prompting Baselines}
\label{subsubsec:edma_prompting_baselines}

Table~\ref{tab:results_mmqa_prompting_combined} summarises the performance of EDMA alongside standard prompting baselines across six representative MLLMs. EDMA consistently outperforms Direct Prompting (DP), Chain-of-Thought (CoT), and Planning Prompting (PP) across all evaluated backbones and datasets. On MultiModalQA, EDMA achieves an average score of $80.1$ EM, $81.1$ F1, and $79.3$ LLM-J, outperforming the strongest prompting baseline (PP) by $+22.7$ EM, $+14.0$ F1, and $+8.5$ LLM-J points. On ManyModalQA, EDMA reaches $83.0$ EM, $83.3$ F1, and $89.1$ LLM-J, improving over PP by $+23.0$ EM, $+11.5$ F1, and $+0.6$ LLM-J. Across both benchmarks, EDMA achieves average gains of $+22.9$ EM, $+12.8$ F1, and $+4.6$ LLM-J over the strongest prompting baseline. The smaller LLM-Judge gain on ManyModalQA (+0.6 vs +8.5 on MultiModalQA) stems from lower benchmark complexity, limiting EDMA's possible improvement space. Nevertheless, EDMA's strong leads in EM (+23.0) and F1 (+11.5) on ManyModalQA confirm that entropy-guided decomposition remains essential for precise entity extraction even on simpler queries.

\begin{table}[h]
\centering
\caption{EDMA achieves the highest average EM, F1, and LLM-judge scores on MultiModalQA, ManyModalQA across all evaluated models, outperforming best-practice prompting strategies: Direct Prompting (DP), Chain-of-Thought (CoT), and Planning Prompting (PP).}
\label{tab:results_mmqa_prompting_combined}

\resizebox{\textwidth}{!}{%
\begin{tabular}{lccccccccccccccccccccc}
\toprule
& \multicolumn{21}{c}{\textbf{Model}} \\
\cmidrule(lr){2-22}
\textbf{Method}
& \multicolumn{3}{c}{\textbf{Claude-4.6}}
& \multicolumn{3}{c}{\textbf{Kimi-K2.5}}
& \multicolumn{3}{c}{\textbf{Qwen3}}
& \multicolumn{3}{c}{\textbf{GPT-5.2}}
& \multicolumn{3}{c}{\textbf{Nemotron}}
& \multicolumn{3}{c}{\textbf{Nova-2-Lite}}
& \multicolumn{3}{c}{\textbf{Avg}} \\
\cmidrule(lr){2-4}
\cmidrule(lr){5-7}
\cmidrule(lr){8-10}
\cmidrule(lr){11-13}
\cmidrule(lr){14-16}
\cmidrule(lr){17-19}
\cmidrule(lr){20-22}
& EM & F1 & LLM-J
& EM & F1 & LLM-J
& EM & F1 & LLM-J
& EM & F1 & LLM-J
& EM & F1 & LLM-J
& EM & F1 & LLM-J
& EM & F1 & LLM-J \\
\midrule

\multicolumn{22}{c}{\textbf{MultiModalQA \cite{talmor2021multimodalqa}}} \\
\midrule

EDMA
& \textbf{82.8} & \textbf{83.5} & \textbf{88.9}
& \textbf{80.4} & \textbf{81.0} & \textbf{83.2}
& \textbf{79.4} & \textbf{80.9} & \textbf{82.5}
& \textbf{80.0} & \textbf{80.3} & \textbf{80.5}
& \textbf{79.0} & \textbf{80.5} & \textbf{75.1}
& \textbf{79.0} & \textbf{80.4} & \textbf{65.9}
& \textbf{80.1} & \textbf{81.1} & \textbf{79.3} \\

PP
& 61.4 & 73.1 & 85.6
& 64.0 & 74.0 & 82.0
& 60.6 & 69.4 & 75.4
& 56.6 & 68.4 & 75.5
& 51.8 & 60.0 & 64.6
& 50.0 & 57.8 & 41.8
& 57.4 & 67.1 & 70.8 \\

CoT
& 61.4 & 72.3 & 87.4
& 63.8 & 72.7 & 79.9
& 61.0 & 70.7 & 76.4
& 46.2 & 56.8 & 64.4
& 46.4 & 54.2 & 57.1
& 43.4 & 50.2 & 53.4
& 53.7 & 62.8 & 69.8 \\

DP
& 60.8 & 70.5 & 82.7
& 39.4 & 47.2 & 52.1
& 60.6 & 68.9 & 74.3
& 58.2 & 68.8 & 75.7
& 42.2 & 50.0 & 52.3
& 51.8 & 59.6 & 64.6
& 52.2 & 60.8 & 66.9 \\

\midrule
\multicolumn{22}{c}{\textbf{ManyModalQA \cite{hannan2020manymodalqa}}} \\
\midrule

EDMA
& \textbf{82.0} & \textbf{82.4} & 90.0
& \textbf{86.0} & \textbf{86.6} & \textbf{94.5}
& \textbf{84.0} & \textbf{84.0} & \textbf{90.0}
& \textbf{88.0} & \textbf{88.5} & \textbf{88.0}
& \textbf{78.0} & \textbf{78.0} & \textbf{84.5}
& \textbf{80.0} & \textbf{80.0} & \textbf{87.5}
& \textbf{83.0} & \textbf{83.3} & \textbf{89.1} \\

PP
& 56.0 & 70.2 & 90.0
& 62.0 & 74.4 & 93.5
& 64.0 & 75.8 & 89.5
& 48.0 & 63.4 & 86.5
& 64.0 & 71.3 & 84.0
& 66.0 & 75.7 & 87.5
& 60.0 & 71.8 & 88.5 \\

DP
& 70.0 & 81.5 & \textbf{93.0}
& 70.0 & 80.1 & 93.5
& 62.0 & 71.2 & 86.0
& 56.0 & 67.8 & 87.5
& 48.0 & 56.7 & 74.5
& 58.0 & 66.7 & 80.5
& 60.7 & 70.7 & 85.8 \\

CoT
& 56.0 & 68.2 & 88.0
& 62.0 & 70.4 & 83.0
& 64.0 & 73.3 & 86.5
& 42.0 & 56.2 & 85.0
& 40.0 & 50.2 & 69.0
& 60.0 & 67.9 & 81.0
& 54.0 & 64.4 & 82.1 \\

\bottomrule
\end{tabular}%
}
\end{table}

In Appendix~\ref{appendix:sec:intermediate_operations} and Appendix~\ref{appendix:sec:statistical_variance}, we respectively report detailed analyses regarding EDMA's intermediate operation accuracy and statistical variance for EDMA and baselines results.

\subsubsection{Comparison between EDMA and Budget Matched Baselines}
\label{subsubsec:edma_budget_matched_baselines}

To determine whether EDMA's gains result merely from increased inference computation, we evaluate EDMA on MultiModalQA against budget-matched self-consistency baselines (CoT-SC and PP-SC with $N=6$ samples) and the multi-agent framework MAMMQA. As shown in Table~\ref{tab:computational_cost} (Appendix~\ref{appendix:sec:access_computational_resources}), EDMA is approximately $6\times$ computationally more expensive than single-pass CoT and PP, making $N=6$ a direct budget-matched baseline. Furthermore, EDMA's computational footprint is comparable to MAMMQA, establishing it as an additional budget-matched baseline. We evaluate MAMMQA using exclusively LLM-J, as its verbose answer style would unfairly penalize it under exact-matching metrics like EM and F1.

\begin{table}[h]
\centering
\caption{Comparison of EDMA with MAMMQA and self-consistency (SC) baselines ($N=6$ samples) across EM, F1, and LLM-J metrics.}
\label{tab:edma_sc}
\resizebox{\textwidth}{!}{%
\begin{tabular}{lccccccccccccccc}
\toprule
& \multicolumn{15}{c}{\textbf{Model}} \\
\cmidrule(lr){2-16}
\textbf{Method}
& \multicolumn{3}{c}{\textbf{Claude-4.6}}
& \multicolumn{3}{c}{\textbf{Kimi-K2.5}}
& \multicolumn{3}{c}{\textbf{Nemotron}}
& \multicolumn{3}{c}{\textbf{Nova-2-Lite}}
& \multicolumn{3}{c}{\textbf{Avg}} \\
\cmidrule(lr){2-4} \cmidrule(lr){5-7} \cmidrule(lr){8-10} \cmidrule(lr){11-13} \cmidrule(lr){14-16}
& EM & F1 & LLM-J & EM & F1 & LLM-J & EM & F1 & LLM-J & EM & F1 & LLM-J & EM & F1 & LLM-J \\
\midrule

EDMA
& \textbf{82.8} & \textbf{83.5} & \textbf{88.9}
& \textbf{80.4} & \textbf{81.0} & 83.2
& \textbf{79.0} & \textbf{80.5} & \textbf{75.1}
& \textbf{79.0} & \textbf{80.4} & \textbf{65.9}
& \textbf{80.3} & \textbf{81.4} & \textbf{78.2} \\

MAMMQA
& -- & -- & 83.3
& -- & -- & \textbf{88.4}
& -- & -- & 65.7
& -- & -- & 65.7
& -- & -- & 75.7 \\

CoT-SC
& 58.5 & 72.0 & 84.3
& 58.5 & 69.6 & 76.4
& 42.5 & 52.7 & 61.0
& 46.5 & 55.6 & 63.2
& 51.5 & 62.5 & 71.2 \\

PP-SC
& 56.5 & 69.9 & 82.5
& 60.5 & 73.7 & 80.8
& 52.5 & 61.6 & 60.2
& 55.5 & 67.3 & 59.6
& 56.3 & 68.1 & 70.8 \\

\bottomrule
\end{tabular}
}
\end{table}

As detailed in Table~\ref{tab:edma_sc}, the effect of self-consistency depends on model capacity. Under PP-SC, Nova-2-Lite gains $+5.5$ EM, $+9.5$ F1, and $+17.8$ LLM-J over single-pass PP, whereas Nemotron improves slightly on EM ($+0.7$) and F1 ($+1.6$) but drops $-4.4$ LLM-J. In contrast, the larger backbones degrade: Claude-4.6 drops $-4.9$ EM, $-3.2$ F1, and $-3.1$ LLM-J, and Kimi-K2.5 drops $-3.5$ EM, $-0.3$ F1, and $-1.2$ LLM-J. Under CoT-SC, Nova-2-Lite again benefits across all metrics ($+3.1$ EM, $+5.4$ F1, $+9.8$ LLM-J over single-pass CoT), while Nemotron gains $+3.9$ LLM-J but loses $-3.9$ EM and $-1.5$ F1. The larger backbones again degrade on all metrics: Claude-4.6 drops $-2.9$ EM, $-0.3$ F1, and $-3.1$ LLM-J, and Kimi-K2.5 drops $-5.3$ EM, $-3.1$ F1, and $-3.5$ LLM-J.

Across the four backbones, SC yields only modest average LLM-J gains over single-pass prompting ($+2.3$ for PP-SC and $+1.8$ for CoT-SC) and does not improve EM. 
In contrast, EDMA ($78.2$ LLM-J, $80.3$ EM, $81.4$ F1) outperforms CoT-SC ($71.2$ LLM-J) and PP-SC ($70.8$ LLM-J) by $+7.0$ and $+7.4$ LLM-J, respectively, and MAMMQA ($75.7$ LLM-J) by $+2.5$ LLM-J on average, although MAMMQA is stronger on Kimi-K2.5 ($88.4$ vs. $83.2$ LLM-J). The importance of EDMA's core mechanisms is further supported by two ablation studies, which isolate the contributions of entropy minimization and cross-modal conditioning (Tables~\ref{tab:ablation_random_trajectory}--\ref{tab:ablation_cross_modal}, Appendix~\ref{appendix:sec:ablation:studies}).

\subsection{Qualitative Example}

\begin{figure}[h]
    \centering
    \includegraphics[width=\linewidth]{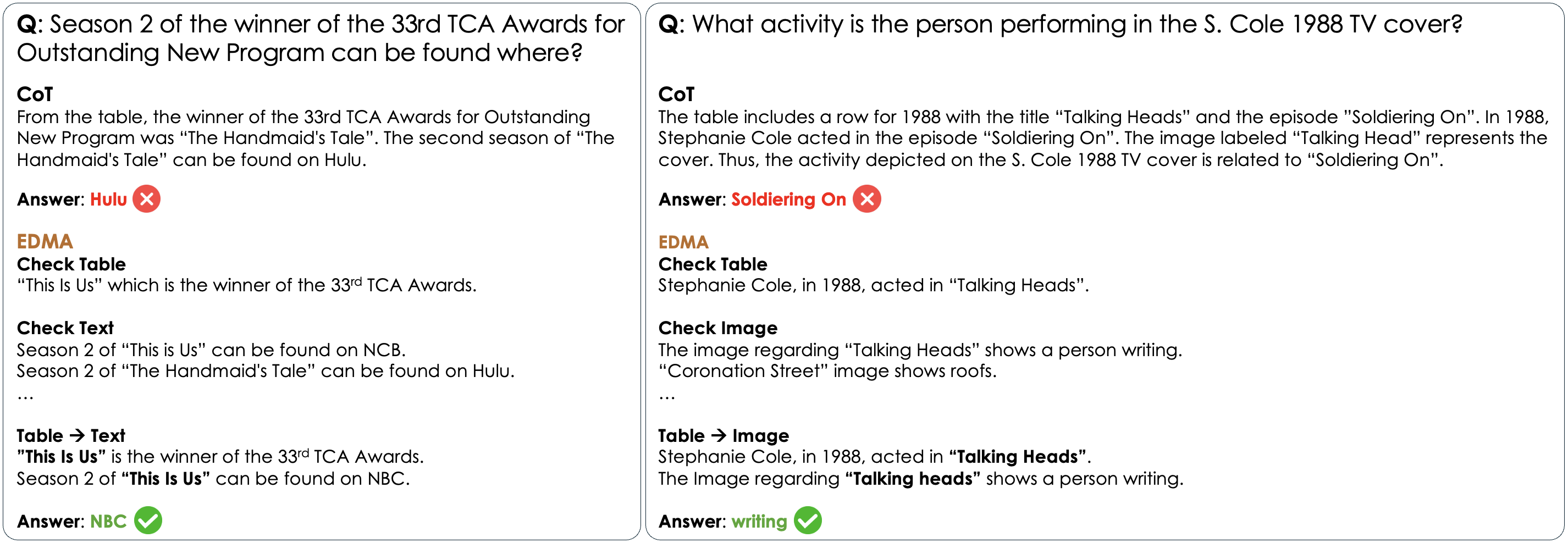}
    \caption{
    Qualitative comparison of CoT and EDMA. EDMA answers both questions correctly, isolating information in the unimodal setting before bridging modalities in the multimodal step, while CoT returns incorrect answers.}
    \label{fig:qualitative_results}
\end{figure}

We attribute the performance gains achieved by EDMA to its staged reasoning process. Specifically, EDMA first decomposes the answer space into unimodal partitions before extracting explicit cross-modal evidence to construct combined multimodal partitions. This progressive refinement allows the model to isolate and verify relevant information within individual modalities prior to cross-modal integration and trajectory selection, which may improve the reliability of evidence aggregation and reduce ambiguity in cross-modal alignment. In contrast, standard prompting paradigms such as CoT and PP depend on the model capacity to implicitly manage decomposition and multimodal integration within a single inference stream, which may limit their ability to consistently isolate and align evidence across modalities.

Figure~\ref{fig:qualitative_results} illustrates this contrast on two multi-hop examples. In the first, CoT misidentifies the winner of the \nth{33} TCA Award for Outstanding New Program. EDMA instead isolates the tabular context to identify ``This Is Us'' as the winner, and then queries the secondary modality to retrieve the correct broadcasting network (NBC). In the second example, CoT retrieves the tabular context (the series ``Talking Heads'') but its prediction is not grounded in the visual modality. EDMA conditions its visual query on the tabular entry and correctly aligns the image, identifying the depicted activity as writing.

\subsection{False Positives Correlate with Higher Entropy}

\begin{figure}[h]
    \begin{subfigure}{0.48\linewidth}
        \centering
        \includegraphics[width=\linewidth]{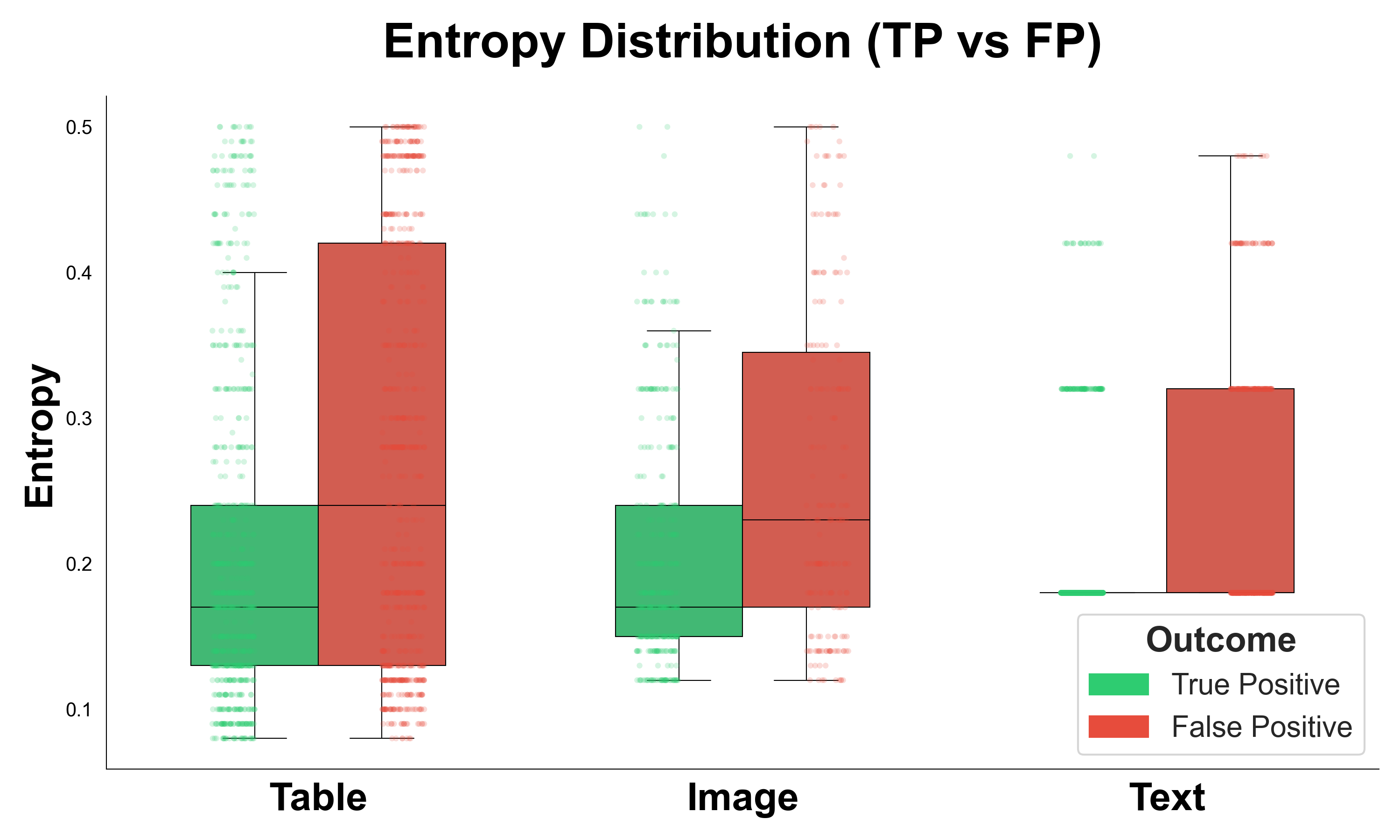}
        \caption{Across all three target modalities, the mean entropy of FPs is higher than for TPs.}
        \label{fig:boxplot_tp_fp}
    \end{subfigure}
    \centering
    \hfill
    \begin{subfigure}{0.48\linewidth}
        \centering
        \includegraphics[width=\linewidth]{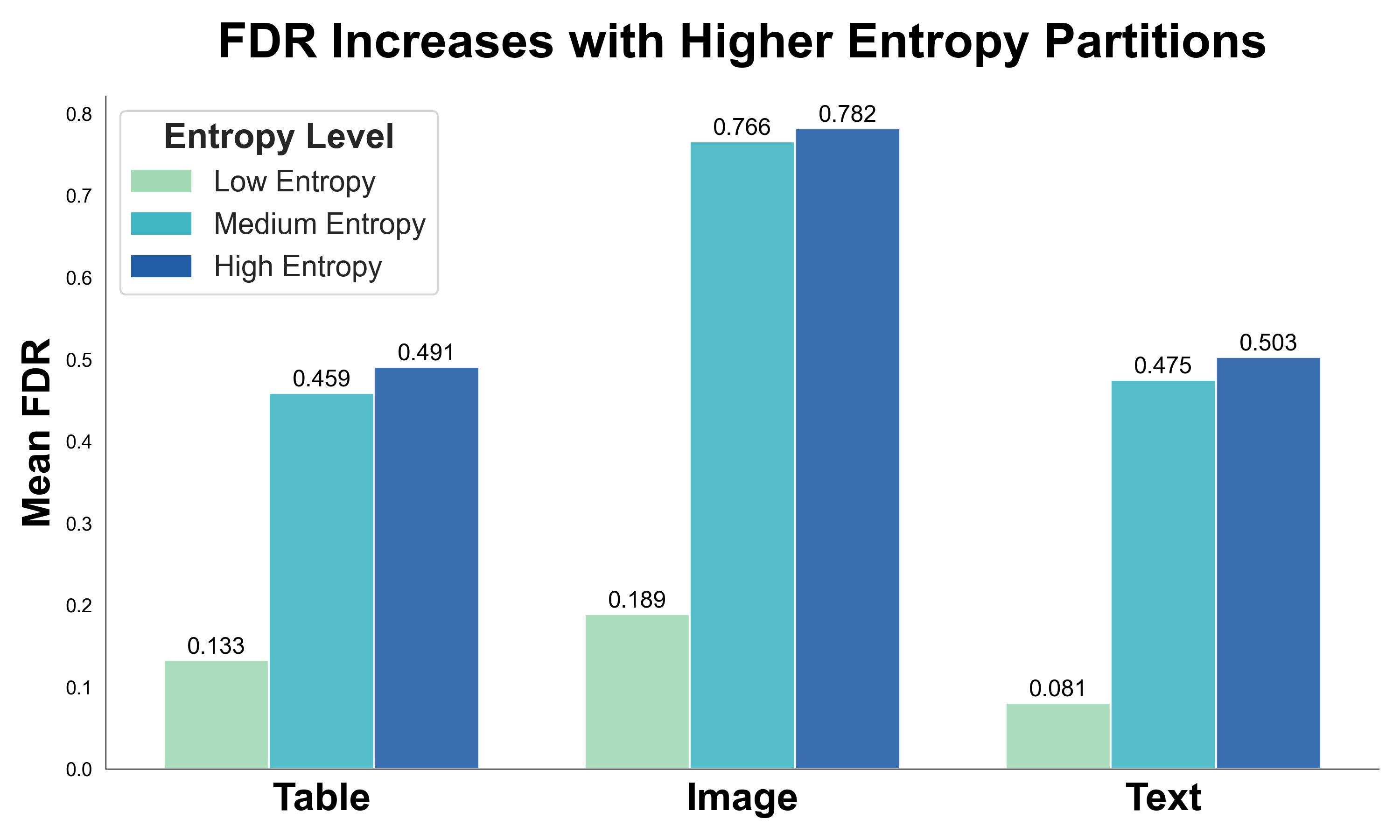}
        \caption{Selecting higher-entropy trajectories is associated with a higher FDR.}
        \label{fig:interventions_graph}
    \end{subfigure}

    \caption{Wrong answers are associated with higher entropy values, as shown by a correlational analysis (a) and a controlled trajectory-selection analysis (b).}
    
    \label{fig:combined_entropy}
\end{figure}

Here we analyse the relationship between entropy and prediction errors to assess the effectiveness of EDMA, examining how entropy correlates with the occurrence of FPs across target modalities. Figure \ref{fig:boxplot_tp_fp} shows that across all modalities, FPs are consistently associated with higher entropy values compared to true positives (TPs). This indicates that incorrect predictions tend to occur in regions of higher uncertainty, where the candidate answer space remains insufficiently constrained. In contrast, correct predictions are more likely to arise when entropy is lower, reflecting a more decisive partitioning of the answer space.
See Table \ref{tab:entropy_analysis_tp_fp} in Appendix \ref{appendix:sec:statistical_significance} for data regarding statistical significance.

\subsection{Answering Trajectories with Lower Entropy Yield Better Answers}
To further show the connection between entropy level and FPs, we perform a controlled intervention analysis to assess whether entropy directly impacts prediction performance. We generate three candidate answers per question by selecting answering trajectories corresponding to the lowest, median, and highest entropy levels. The low-entropy trajectory corresponds to the one selected by EDMA in the main experiments. This design keeps the input question fixed while varying the selected trajectory, and hence its entropy. We evaluate prediction reliability using the False Discovery Rate (FDR), defined as the proportion of FPs among all positive predictions. We compute FDR separately for each entropy group across modalities. Figure \ref{fig:interventions_graph} shows a consistent monotonic trend: low-entropy trajectories yield the lowest FDR, while medium- and high-entropy trajectories exhibit progressively higher error rates. This pattern is consistent across all target modalities, indicating a strong association between higher logical entropy and reduced prediction reliability. By systematically varying the selected trajectory while keeping the input query fixed, we observe a robust degradation in performance as entropy increases. These results support the importance of entropy minimization in EDMA: lower-entropy trajectories are associated with fewer false positives and improved prediction performance. Tables \ref{tab:friedman_global}--\ref{tab:pairwise_wilcoxon} in Appendix \ref{appendix:sec:statistical_significance}, report data regarding statistical significance.

\section{Related work}

\paragraph{MultimodalQA} 
Multimodal question answering (MMQA) has been introduced as a challenging task requiring the integration of heterogeneous data sources such as text, tables, and images \cite{talmor2021multimodalqa}. Subsequent work has focused on improving representation and integration across modalities, including unified architectures such as UniMMQA \cite{luo2023unifying}, Solar \cite{yu2023unified}, retrieval-augmented approaches such as UniRAG \cite{lim2024unirag}, and conversational extensions like MMConvQA \cite{li2022mmcoqa}. More recent approaches leverage LLMs and MLLMs to enhance multi-step inference capabilities, for instance by combining language and vision models \cite{liu2025union}, improving cross-modal alignment through similarity computation \cite{zhang2025enhancing}, or framing information synthesis as a multi-agent process, as in MAMMQA \cite{anvekar2025rethinking}.

\paragraph{LLMs in MultimodalQA}
LLMs have been widely adopted to enhance multimodal QA, often through extensions of Chain-of-Thought (CoT) prompting. Prior work includes Chain-of-Action (CoA) \cite{pan2024chain}, knowledge-augmented CoT (KAM-CoT) \cite{mondal2024kam}, duty-distinct CoT (DDCoT) \cite{zheng2023ddcot}, and frameworks for structured CoT such as UniCoTT \cite{zhuang2025unicott}. These approaches improve inference by decomposing complex tasks into intermediate steps or incorporating external knowledge. Additional adaptations target specific modalities, such as AVQA-CoT for audio-visual QA \cite{li2024avqa} and methods that leverage textual signals to guide multimodal understanding \cite{wang2024can}. Despite these advances, existing CoT-based approaches do not provide explicit mechanisms to quantify the extent to which each modality contributes to the final prediction.

\paragraph{Uncertainty as Guidance}
Recent work has explored incorporating uncertainty quantification into LLM reasoning to improve robustness and decision-making. Hu et al. \cite{hu2024uncertainty} propose Uncertainty of Thoughts, which evaluates alternative planning paths based on their expected uncertainty reduction. Yin et al. \cite{yin2024reasoning} introduce uncertainty-aware adaptive guidance, an approach that relies on uncertainty signals from the LLM to guide the reasoning process. Liang et al. \cite{liang2024introspective} propose introspective planning to align model uncertainty with task ambiguity, while Bayat et al. \cite{bayat2024enhanced} leverage uncertainty estimates to guide answer selection. While these approaches treat uncertainty as a global signal for improving reasoning, they do not provide fine-grained, modality-level attribution of uncertainty reduction, limiting their effectiveness in multimodal QA settings.

\section{Limitations}
\label{sec:limitations}
EDMA's worst-case complexity is $O(n^3)$ when modalities, requirements, and elements grow together. In practice this translates into many LLM calls per question, costing between $5$ and $6.4\times$ more and taking between $20$ and $46\times$ longer than the prompting baselines (Appendix~\ref{appendix:sec:access_computational_resources}). Our budget-matched baselines match its monetary cost, but not its latency or number of calls, and a full accuracy--cost analysis is left to future work.

Logical entropy is computed under a uniform distribution over elements, which avoids calibrated probabilities but ignores differences in the relevance of individual elements.

Finally, our methodology does not explicitly address scenarios in which different modalities provide conflicting information. One possible extension would be to partition the answer space based on parallel interpretations of the available evidence, treating them as distinct answering trajectories, which would let the model explore potentially conflicting paths and evaluate their outcomes independently.


\section{Conclusion}
This paper introduces Entropy-Driven Multimodal Answering (EDMA), a novel methodology for multimodal QA based on logical entropy minimization and the construction of quantifiable answering trajectories. 

By measuring how each step reduces logical entropy, EDMA quantifies the contribution of every modality, improving traceability. Our analyses show that false positives are associated with higher entropy, and that choosing the lowest-entropy trajectory tends to lead to better answers, making entropy a useful proxy for performance. Overall, on the same backbones, EDMA improves over the baselines it is compared against.

\section*{Acknowledgments and Disclosures}
This work was partially supported by the \textit{Nederlandse Organisatie voor Wetenschappelijk Onderzoek (NWO)} under the KIC HEWSTI Project under grant no. KIC1.VE01.20.004. 

The authors declare that they have no competing financial interests or personal relationships that could have appeared to influence the work reported in this paper.

\bibliographystyle{unsrt}
\bibliography{references}



\appendix

\newpage
\section{Notation}

\begin{table}[h]
\centering
\caption{Summary of notation used.}
\begin{tabular}{ll}
\toprule
\textbf{Symbol} & \textbf{Description} \\
\midrule
$\question$ & Input question \\
$\modalities$ & Set of all modalities \\
$\modality \in \modalities$ & A specific modality \\
$\startmodalities \subseteq \modalities$ & Set of selected starting modalities \\
$\analysemodalites$ & Modalities under analysis \\
$\targetmodality$ & Candidate target modality \\
$\conditioningmodalities$ & Set of conditioning modalities ($\analysemodalites \setminus \{\targetmodality\}$) \\
$\elements$ & Set of elements for modality $\modality$ \\
$\element \in \elements$ & A specific element of modality $\modality$ \\
$\requirements$ & Set of answer requirements \\
$\requirement \in \requirements$ & A specific answer requirement \\
$\omegafilling$ & Elements of modality $\modality$ satisfying requirement $\requirement$ \\
$\omeganotfilling$ & Elements of modality $\modality$ not satisfying $\requirement$ \\
$\unipartition$ & Partition of modality $\modality$ w.r.t. requirement $\requirement$ \\
$\unientropy$ & Logical entropy of partition $\unipartition$ \\
$\mathcal{H}_{uni}[\modality, \requirement]$ & Cache storing $(\unientropy, \omegafilling)$ for each $(\modality, \requirement)$ pair \\
$\partialanswers$ & Set of cross-modal evidence extracted from a modality \\
$E[\modality, \requirement]$ & Cached cross-modal evidence extracted from modality $\modality$ w.r.t. requirement $\requirement$ \\
$\omegatarget$ & Target elements consistent with requirement $\requirement$ and cross-modal evidence $\partialanswers$ \\
$\omegatargetneg$ & Target elements consistent with $\requirement$ but inconsistent with $\partialanswers$ \\
$\targetpartition$ & Conditional partition of target modality given $\partialanswers$ \\
$\targetentropy$ & Multimodal logical entropy after applying $\partialanswers$ \\
$\Omega_{curr}$ & Elements of $\targetmodality$ still consistent with $\requirement$, refined across conditioning steps \\
$E_{curr}$ & Cross-modal evidence accumulated over the conditioning steps \\
$\mathcal{H}_{multi}[\targetmodality, \requirement]$ & Cache storing $(\targetentropy, \Omega_{curr}, E_{curr})$ for each $(\targetmodality, \requirement)$ pair \\
$\deltaentropy$ & Relative change between $H(P_{m_t}^r)$ and $\targetentropy$ \\
$\trajectoryset$ & Set of candidate unimodal and multimodal trajectories \\
$\optimaltrajectory$ & Optimal trajectory selected via entropy minimization \\
$\finalelements$ & Subset of elements in $\modality^*$ consistent with $\requirement^*$ and $\finalpartialanswers$, as selected by $\optimaltrajectory$ \\
$\finalpartialanswers$ & Cross-modal evidence used for answer generation \\
$\answer$ & The final generated answer \\
\bottomrule
\end{tabular}
\label{tab:notation}
\end{table}

\newpage
\section{Algorithm}
\label{appendix:sec:algorithm}
The EDMA algorithm has a worst-case computational complexity of $O(M \cdot R \cdot K + M^2 \cdot R)$, where $M$ denotes the number of modalities, $R$ the number of requirements, and $K$ the maximum number of elements per modality. Stage 1 (Unimodal Partitioning) contributes $O(M \cdot R \cdot K)$, evaluating each of the $K$ elements per modality against each requirement, and also caches the cross-modal evidence extracted for each $(m, r)$ pair. Stage 2 (Cross-Modal Sequential Conditioning) contributes $O(M^2 \cdot R)$, refining each of the $M \cdot R$ target pairs by conditioning on the remaining $M$ modalities via $O(1)$ lookups into this cache. Under $M \sim R \sim K \sim n$, the overall complexity is $O(n^3)$.

\begin{algorithm}
\caption{Entropy-Driven Multimodal Answering (EDMA)}
\label{alg:edma}
\begin{algorithmic}[1]
\Require $\question$, $\modalities$, $\{\elements, \modality \in \modalities\}$
\Ensure Final answer $\answer$

\State $\requirements \gets LLM(\prompt_{\requirements}, \question)$ \hfill \textit{// Extraction of answer requirements}
\State $\startmodalities \gets LLM(\prompt_{\startmodalities}, \question, \{\elements, \modality \in \modalities\})$ \hfill \textit{// Prediction of Starting Modalities}
\State $\answer \gets \text{None}$
\State $\analysemodalites \gets \emptyset$

\While{$\answer = \text{None}$ \textbf{and} $\startmodalities \neq \emptyset$}
    \State $\analysemodalites \gets \analysemodalites \cup \startmodalities$
    \State $\modalities \gets \modalities \setminus \startmodalities$
    
    \Statex \hfill \textit{// Stage 1: Unimodal Partitioning (with Extraction Caching)}
    \For{$(\modality, \requirement) \in \analysemodalites \times \requirements$}
        \State Generate $\unipartition = \{\omegafilling, \omeganotfilling\}$ and compute $\unientropy$
        \State Store $\unientropy$ and $\omegafilling$ in $\mathcal{H}_{uni}[\modality, \requirement]$
        \State $E[\modality, \requirement] \gets \{LLM(\prompt_{Extract}, \question, \element) \mid \element \in \omegafilling\}$ \hfill \textit{// Cached once per $(m,r)$}
    \EndFor

    \Statex \hfill \textit{// Stage 2: Cross-Modal Sequential Conditioning}
    \For{$(\targetmodality, \requirement) \in \analysemodalites \times \requirements$}
        \State $\conditioningmodalities \gets \analysemodalites \setminus \{\targetmodality\}, \, \Omega_{curr} \gets \omegafillingtarget, \, E_{curr} \gets \emptyset$
        
        \For{$\modality \in \conditioningmodalities$}
            \State $\partialanswers \gets E[\modality, \requirement]$ \hfill \textit{// $O(1)$ lookup}
            \State Generate $\targetpartition = \{\omegatarget, \omegatargetneg\}$ using $\Omega_{curr}$ and $\partialanswers$
            \State $\Omega_{curr} \gets \omegatarget$
            \State $E_{curr} \gets E_{curr} \cup \partialanswers$
        \EndFor
        \State $\targetentropy \gets H(\targetpartition)$
        \State Store $\targetentropy, \Omega_{curr}, E_{curr}$ in $\mathcal{H}_{multi}[\targetmodality, \requirement]$
    \EndFor

    \Statex \hfill \textit{// Stage 3: Trajectory Selection}
    \State $\mathcal{T} \gets \emptyset$ 
    \For{$(\targetmodality, \requirement) \in \analysemodalites \times \requirements$}
        \State $H_{u}, \Omega_{u} \gets \mathcal{H}_{uni}[\targetmodality, \requirement]$
        \State $H_{m}, \Omega_{m}, E_{m} \gets \mathcal{H}_{multi}[\targetmodality, \requirement]$
        
        \State $\mathcal{T} \gets \mathcal{T} \cup \{(\targetmodality, \requirement, H_{u}, \Omega_{u}, \emptyset)\}$ \hfill \textit{// Unimodal path}
        
        \State \textbf{if} $H_{u} > 0$ \textbf{then} $\deltaentropy \gets (H_{u} - H_{m}) / H_{u}$ \textbf{else} $\deltaentropy \gets 0$ 
        \State $\mathcal{T} \gets \mathcal{T} \cup \{(\targetmodality, \requirement, H_{m}, \Omega_{m}, E_{m}, \deltaentropy)\}$ \hfill \textit{// Multimodal path}
    \EndFor

    \State Select $t^* = (\modality^*, \requirement^*, H^*, \finalelements, E^*)$ from $\mathcal{T}$ that minimizes $H$

    \Statex \hfill \textit{// Stage 4: Final Inference}
    \State $\answer \gets LLM(\question, \finalelements, E^*)$ \hfill \textit{// Inference from optimal subset and evidence}
    
    \If{$\answer = \text{None}$ \textbf{and} $\modalities \neq \emptyset$}
        \State $\startmodalities \gets LLM(\prompt_{\startmodalities}, \question, \{\elements, \modality \in \modalities\})$
    \Else
        \State $\startmodalities \gets \emptyset$
    \EndIf
\EndWhile
\State \Return $\answer$
\end{algorithmic}
\end{algorithm}

\section{Experimental Setup}
\label{appendix:sec:experimentalsetups}

\subsection{Baselines}
\label{appendix:subsec:baselines}

In this section, we provide detailed descriptions of the baselines used in our experiments. We categorize these into multimodal architectures and prompting-based strategies.

\subsubsection{Multimodal QA Systems}

\paragraph{MAMMQA \cite{anvekar2025rethinking}} A multi-agent QA framework for multimodal inputs spanning text, tables, and images. The system comprises two Vision-Language Model (VLM) agents and one text-based LLM agent. The first VLM decomposes the query into sub-questions to sequentially retrieve partial evidence across modalities, the second VLM synthesizes intermediate insights via cross-modal reasoning, and the LLM aggregates the findings into a cohesive final answer.

\paragraph{MultiModalQA \cite{talmor2021multimodalqa}} The baseline pipeline introduced alongside the MultiModalQA benchmark. It employs modality-specific information extractors combined with an \textit{ImplicitDecomp} module to perform multi-step question decomposition and answer synthesis across text, tables, and images.

\paragraph{UniMMQA \cite{luo2023unifying}} A unified multimodal QA framework designed to reduce the modality gap by projecting heterogeneous data (text, tables, and images) into a shared representation space, enabling integrated cross-modal retrieval and answer generation.

\paragraph{Solar \cite{yu2023unified}} A sequence-to-sequence multimodal framework that unifies visual and textual token processing within a single transformer architecture, optimized for cross-modal alignment through joint pre-training objectives.

\paragraph{UniRAG \cite{lim2024unirag}} A retrieval-augmented generation (RAG) framework that unifies evidence retrieval across textual, visual, and tabular corpora before feeding the aggregated context into a Large Multimodal Model (LMM) for answer generation.

\paragraph{ManyModalQA \cite{hannan2020manymodalqa}} The baseline framework introduced with the ManyModalQA benchmark. It incorporates an explicit modality routing mechanism to determine the most relevant data source (text, image, or table) prior to applying specialized extraction modules to return the final answer.

\subsubsection{Prompting Strategies}

\paragraph{Direct Prompting (DP)} A standard zero-shot prompting approach where the model receives the input context and query, and directly outputs the final answer without explicit intermediate reasoning steps.

\paragraph{Chain-of-Thought (CoT) \cite{wei2022chain}} A prompting method that prompts the model to generate a sequence of intermediate natural language reasoning steps before outputting the final answer within a single generation stream.

\paragraph{Planning Prompting (PP) \cite{wang2023plan}} A two-stage prompting technique that decouples reasoning planning from execution: the model first generates a high-level plan of sub-tasks, and subsequently executes each sub-task to derive the final answer.
    
\paragraph{CoT Self-Consistency (CoT-SC) \cite{wang2022self}} An ensemble strategy that samples $N$ independent Chain-of-Thought reasoning paths and selects the final answer via majority voting.

\paragraph{PP Self-Consistency (PP-SC)} An extension of self-consistency \cite{wang2022self} applied to Planning Prompting, sampling $N$ independent plan-and-execute reasoning trajectories and selecting the answer via majority voting.

\subsection{Datasets}

\textbf{MultiModalQA} \cite{talmor2021multimodalqa} is a comprehensive benchmark designed to evaluate a model's ability to perform joint integration across three distinct data modalities: structured tables, natural language text, and images. Unlike earlier multimodal datasets where the answer usually resides within a single source (e.g., just an image or just a paragraph), MultiModalQA is specifically engineered to require multi-hop reasoning. We report results on the validation set since ground-truth annotations for the test set are not publicly accessible.

To further show the generalizability of our methodology, we test it also on \textbf{ManyModalQA} \cite{hannan2020manymodalqa}, which is a multimodal question-answering benchmark designed to evaluate an agent's ability to reason across three distinct data types: textual passages, structured tables, and visual images. Unlike standard multimodal datasets where the target modality is often implied by the question, ManyModalQA contains ambiguous questions where the source of the answer cannot be determined by the question text alone.



\section{Comparison with Previous Multimodal QA Systems}
\label{appendix:sec:additional_results}
Table \ref{tab:results_mmqa_stacked} compares EDMA with prior multimodal QA systems. EDMA achieves the best performance across both datasets. On MultiModalQA, it obtains $82.8$ EM and $83.5$ F1, improving over the strongest prior system (UniMMQA) by $+9.1$ EM and $+3.4$ F1. On ManyModalQA, the gains are substantially larger, with improvements of $+35.9$ EM and $+36.1$ F1 over UniMMQA. We note that these comparisons reflect both methodological improvements and differences in underlying foundation models.

\begin{table}[h]
\centering
\caption{Comparison of EDMA against prior specialized multimodal QA frameworks on MultiModalQA and ManyModalQA datasets.}
\label{tab:results_mmqa_stacked}

\resizebox{0.70\textwidth}{!}{%
\begin{tabular}{llcc}
\toprule
\textbf{Approach} & \textbf{Model} & \textbf{EM} & \textbf{F1} \\
\midrule

\multicolumn{4}{c}{\textbf{MultiModalQA \cite{talmor2021multimodalqa}}} \\
\midrule

EDMA 
& Claude-4.6
& \textbf{82.8} & \textbf{83.5} \\

UniMMQA \cite{luo2023unifying}      
& T5-3B \cite{raffel2020exploring}                       
& 73.7 & 80.1 \\

UniRAG \cite{lim2024unirag}       
& LLaVA-v1.5-7B \cite{liu2023visual}           
& 67.4 & 71.3 \\

Solar \cite{yu2023unified}        
& T5 \cite{raffel2020exploring} + BLIP \cite{li2022blip}             
& 59.8 & 66.1 \\

MultiModalQA \cite{talmor2021multimodalqa} 
& RoBERTa-large \cite{liu2019roberta} + ViLBERT \cite{lu2019vilbert} 
& 49.3 & 55.9 \\

\midrule

\multicolumn{4}{c}{\textbf{ManyModalQA \cite{hannan2020manymodalqa}}} \\
\midrule

EDMA
& GPT-5.2
& \textbf{88.0} & \textbf{88.5} \\

UniMMQA \cite{luo2023unifying} 
& T5-3B \cite{raffel2020exploring}   
& 52.1 & 52.4 \\

ManyModalQA \cite{hannan2020manymodalqa} 
& ELMo \cite{peters-etal-2018-deep} + BERT \cite{devlin2019bert} 
& 39.7 & 39.7 \\

\bottomrule
\end{tabular}%
}
\end{table}

\section{Models}
\label{appendix:sec:access_computational_resources}

\subsection{Access to Models}
We access GPT-5.2 via its official API and all remaining models through Amazon Bedrock. For all experiments, we use greedy decoding ($\text{temperature} = 0$) with a maximum generation length of $1000$ tokens.

\subsection{Computational Resources}
Table \ref{tab:computational_cost} details the raw average token consumption, API call counts, monetary cost, and latency per model across evaluated approaches (Nova-2-Lite, GPT-5.2, Kimi-K2.5, Nemotron, Qwen3, and Claude-4.6). Furthermore, Table \ref{tab:edma_overhead} summarizes the relative computational overhead of EDMA compared to each baseline strategy.

\begin{table}[h]
\centering
\caption{Average computational cost and latency across models and approaches.}
\label{tab:computational_cost}
\resizebox{\textwidth}{!}{%
\begin{tabular}{llrrrr}
\toprule
\textbf{Model} & \textbf{Approach} & \textbf{Avg. Total Tokens} & \textbf{Avg. \# Calls} & \textbf{Avg. Cost (\$)} & \textbf{Latency (s)} \\
\midrule
Nova-2-Lite & CoT    & 19,775.20 & 1.00   & 0.006 & $\sim$5.04 \\
Nova-2-Lite & DP     & 19,576.12 & 1.00   & 0.006 & $\sim$3 \\
Nova-2-Lite & EDMA   & 89,909.88 & 143.35 & 0.027 & $\sim$30 \\
Nova-2-Lite & PP     & 20,070.06 & 1.00   & 0.007 & $\sim$5 \\
Nova-2-Lite & MAMMQA & 79,972.14 & 7.00   & 0.024 & $\sim$10 \\
\midrule
GPT-5.2 & CoT  & 4,616.74  & 1.00   & 0.009 & $\sim$3 \\
GPT-5.2 & DP   & 4,511.48  & 1.00   & 0.008 & $\sim$1.50 \\
GPT-5.2 & EDMA & 68,574.99 & 181.75 & 0.120 & $\sim$120 \\
GPT-5.2 & PP   & 4,935.78  & 1.00   & 0.013 & $\sim$2.75 \\
\midrule
Kimi-K2.5 & CoT    & 19,812.28 & 1.00   & 0.012 & $\sim$2 \\
Kimi-K2.5 & DP     & 19,587.54 & 1.00   & 0.012 & $\sim$3.5 \\
Kimi-K2.5 & EDMA   & 91,275.13 & 165.43 & 0.055 & $\sim$90 \\
Kimi-K2.5 & PP     & 20,021.98 & 1.00   & 0.013 & $\sim$4.07 \\
Kimi-K2.5 & MAMMQA & 80,172.47 & 7.00   & 0.048 & $\sim$11 \\
\midrule
Nemotron & CoT    & 19,751.52 & 1.00   & 0.003 & $\sim$3 \\
Nemotron & DP     & 19,602.88 & 1.00   & 0.003 & $\sim$1 \\
Nemotron & EDMA   & 90,438.65 & 151.03 & 0.014 & $\sim$90 \\
Nemotron & PP     & 19,828.78 & 1.00   & 0.003 & $\sim$3 \\
Nemotron & MAMMQA & 79,951.83 & 7.00   & 0.012 & $\sim$5 \\
\midrule
Qwen3 & CoT  & 19,836.08 & 1.00   & 0.005 & $\sim$20 \\
Qwen3 & DP   & 19,574.92 & 1.00   & 0.004 & $\sim$5 \\
Qwen3 & EDMA & 90,314.85 & 149.31 & 0.020 & $\sim$120 \\
Qwen3 & PP   & 20,047.36 & 1.00   & 0.005 & $\sim$18.37 \\
\midrule
Claude-4.6 & CoT    & 19,871.12 & 1.00   & 0.063 & $\sim$18 \\
Claude-4.6 & DP     & 19,586.50 & 1.00   & 0.059 & $\sim$4 \\
Claude-4.6 & EDMA   & 91,322.96 & 165.35 & 0.274 & $\sim$180 \\
Claude-4.6 & PP     & 19,937.36 & 1.00   & 0.063 & $\sim$16.01 \\
Claude-4.6 & MAMMQA & 81,253.41 & 7.00   & 0.244 & $\sim$12 \\
\bottomrule
\end{tabular}
}
\end{table}

As shown in Table \ref{tab:edma_overhead}, EDMA is on average $22.83\times$ slower than CoT, $45.79\times$ slower than DP, $19.92\times$ slower than PP, and $11.05\times$ slower than MAMMQA. This reflects the higher computational complexity of the algorithm. Regarding financial cost, EDMA is on average $5.06\times$ more expensive than PP, $5.91\times$ more expensive than CoT, and $6.40\times$ more expensive than DP. However, its cost is directly comparable to MAMMQA ($1.14\times$). Although MAMMQA executes only seven API calls per query, it repeatedly inputs all data modalities (text, images, and tables) into the LLM context, making its token consumption and cost comparable to EDMA. Note that estimates for MAMMQA were computed across all models except GPT-5.2 and Qwen3, where evaluations were omitted due to budget restrictions.

\begin{table}[h]
\centering
\caption{Average cost and latency overhead of EDMA compared with other approaches.}
\label{tab:edma_overhead}
\resizebox{\linewidth}{!}{%
\begin{tabular}{llll}
\toprule
\textbf{Approach} &
\textbf{Avg. Absolute Cost Difference vs. EDMA (\$)} &
\textbf{Avg. Cost Ratio ($\frac{\text{EDMA}}{\text{Approach}}$)} &
\textbf{Avg. Latency Ratio ($\frac{\text{EDMA}}{\text{Approach}}$)} \\
\midrule
CoT    & 0.069 & $5.91\times$ & $22.83\times$ \\
DP     & 0.070 & $6.40\times$ & $45.79\times$ \\
PP     & 0.068 & $5.06\times$ & $19.92\times$ \\
MAMMQA & 0.011 & $1.14\times$ & $11.05\times$ \\
\bottomrule
\end{tabular}%
}
\end{table}


\section{Analysis of Intermediate Operations}
\label{appendix:sec:intermediate_operations}
Table~\ref{tab:error_analysis} reports the average accuracy across each intermediate phase of EDMA, consisting in modality selection, criteria extraction, intermediate answer extraction, and multimodal fusion, evaluated specifically on questions where EDMA produced an incorrect final answer.

\begin{table}[h]
\centering
\caption{Intermediate-step accuracy for questions incorrectly answered by EDMA.}
\label{tab:error_analysis}
\resizebox{0.65\textwidth}{!}{%
\begin{tabular}{lcccc}
\toprule
\textbf{Model} & \textbf{\begin{tabular}[c]{@{}c@{}}Modality\\ Selection\end{tabular}} & \textbf{\begin{tabular}[c]{@{}c@{}}Criteria\\ Extraction\end{tabular}} & \textbf{\begin{tabular}[c]{@{}c@{}}Intermediate\\ Answers\end{tabular}} & \textbf{\begin{tabular}[c]{@{}c@{}}Multimodal\\ Fusion\end{tabular}} \\
\midrule
Nova-2-Lite & 90\% & 94\% & 92\% & 4\% \\
GPT-5.2            & 96\% & 98\% & 96\% & 6\% \\
Kimi-K2.5          & 96\% & 96\% & 96\% & 8\% \\
Nemotron  & 90\% & 94\% & 30\% & 2\% \\
Qwen3      & 96\% & 96\% & 96\% & 8\% \\
Claude-4.6  & 96\% & 98\% & 96\% & 8\% \\
\bottomrule
\end{tabular}
}
\end{table}

Across most models, errors primarily originate during multimodal fusion, where the system must cross-reference information extracted from one modality (e.g., text) with another (e.g., images). In these cases, incorrect final answers stem from over-filtered or wrongly filtered partitions, resulting in the selection of a low-entropy yet incorrect multimodal partition. Nemotron is the sole exception: its failures are driven by low accuracy at the earlier Intermediate Answers stage ($30\%$), causing trajectories to fail before reaching the fusion step.

Table \ref{tab:intermediate_accuracy} presents the corresponding accuracy metrics across all evaluated questions, both correctly and incorrectly answered. Early-stage accuracies (modality selection, criteria extraction, and intermediate answer extraction) remain consistently high across the full dataset. The critical divergence appears in the multimodal fusion phase, confirming that cross-modal fusion constitutes the main point of failure in the methodology.

\begin{table}[h]
\centering
\caption{Intermediate-step accuracy across all questions, both correctly and incorrectly answered.}
\label{tab:intermediate_accuracy}
\resizebox{0.65\textwidth}{!}{%
\begin{tabular}{lcccc}
\toprule
\textbf{Model} & \textbf{\begin{tabular}[c]{@{}c@{}}Modality\\ Selection\end{tabular}} & \textbf{\begin{tabular}[c]{@{}c@{}}Criteria\\ Extraction\end{tabular}} & \textbf{\begin{tabular}[c]{@{}c@{}}Intermediate\\ Answers\end{tabular}} & \textbf{\begin{tabular}[c]{@{}c@{}}Multimodal\\ Fusion\end{tabular}} \\
\midrule
Nova-2-Lite & 95\% & 96\% & 93\% & 70\% \\
GPT-5.2            & 98\% & 98\% & 98\% & 82\% \\
Kimi-K2.5          & 98\% & 97\% & 98\% & 86\% \\
Nemotron  & 93\% & 94\% & 89\% & 76\% \\
Qwen3      & 97\% & 96\% & 97\% & 85\% \\
Claude-4.6  & 98\% & 98\% & 97\% & 91\% \\
\bottomrule
\end{tabular}
}
\end{table}

\section{EDMA Statistical Variance}
\label{appendix:sec:statistical_variance}
Table \ref{tab:determinism_results} reports the mean and standard deviation calculated across ten independent runs. The analysis confirms that the performance advantages observed in single-point comparisons remain stable under repeated evaluation: EDMA's lead over MAMMQA on Claude-4.6, Nemotron, and Nova-2-Lite, as well as MAMMQA's advantage on Kimi-K2.5, hold consistently across runs.

\begin{table}[h]
\centering
\caption{Mean $\pm$ standard deviation of EM, F1, and LLM-judge score (LLM-J) over 10 runs at temperature 0. MAMMQA is only evaluated with LLM-J.}
\label{tab:determinism_results}
\resizebox{0.80\textwidth}{!}{%
\begin{tabular}{llccccc}
\toprule
\textbf{Model} & \textbf{Metric} & \textbf{EDMA} & \textbf{PP} & \textbf{DP} & \textbf{CoT} & \textbf{MAMMQA} \\
\midrule

\multirow{3}{*}{Claude-4.6}
& EM    & \textbf{82.7 $\pm$ 1.9} & 57.9 $\pm$ 1.9 & 56.1 $\pm$ 2.5 & 57.8 $\pm$ 2.0 & -- \\
& F1    & \textbf{83.0 $\pm$ 1.1} & 70.7 $\pm$ 1.3 & 68.1 $\pm$ 1.3 & 70.8 $\pm$ 0.8 & -- \\
& LLM-J & \textbf{88.8 $\pm$ 0.5} & 85.4 $\pm$ 0.4 & 82.6 $\pm$ 0.5 & 87.5 $\pm$ 0.5 & 82.6 $\pm$ 1.3 \\
\midrule

\multirow{3}{*}{Kimi-K2.5}
& EM    & \textbf{80.7 $\pm$ 1.8} & 60.2 $\pm$ 2.1 & 61.9 $\pm$ 0.6 & 58.1 $\pm$ 3.1 & -- \\
& F1    & \textbf{81.3 $\pm$ 0.3} & 72.4 $\pm$ 1.0 & 73.7 $\pm$ 0.3 & 68.7 $\pm$ 2.3 & -- \\
& LLM-J & 83.3 $\pm$ 0.5 & 81.7 $\pm$ 0.4 & 52.0 $\pm$ 0.4 & 79.3 $\pm$ 0.4 & \textbf{88.3 $\pm$ 0.9} \\
\midrule

\multirow{3}{*}{Nemotron}
& EM    & \textbf{80.0 $\pm$ 0.8} & 51.9 $\pm$ 0.9 & 41.2 $\pm$ 0.7 & 46.4 $\pm$ 0.2 & -- \\
& F1    & \textbf{80.7 $\pm$ 0.8} & 61.3 $\pm$ 0.9 & 50.1 $\pm$ 0.2 & 54.8 $\pm$ 0.4 & -- \\
& LLM-J & \textbf{74.9 $\pm$ 0.6} & 64.9 $\pm$ 0.6 & 52.2 $\pm$ 0.4 & 57.0 $\pm$ 0.5 & 66.2 $\pm$ 1.1 \\
\midrule

\multirow{3}{*}{Nova-2-Lite}
& EM    & \textbf{79.1 $\pm$ 0.6} & 47.2 $\pm$ 1.8 & 51.1 $\pm$ 0.6 & 42.3 $\pm$ 0.6 & -- \\
& F1    & \textbf{80.4 $\pm$ 1.1} & 57.1 $\pm$ 1.5 & 60.3 $\pm$ 0.7 & 50.6 $\pm$ 0.3 & -- \\
& LLM-J & \textbf{66.6 $\pm$ 0.4} & 42.1 $\pm$ 0.3 & 64.4 $\pm$ 0.4 & 53.3 $\pm$ 0.3 & 65.9 $\pm$ 1.4 \\
\midrule

\multirow{3}{*}{Average}
& EM    & \textbf{80.7 $\pm$ 1.2} & 54.3 $\pm$ 1.7 & 52.6 $\pm$ 1.1 & 51.2 $\pm$ 1.5 & -- \\
& F1    & \textbf{81.3 $\pm$ 0.8} & 65.4 $\pm$ 1.2 & 63.1 $\pm$ 0.6 & 61.2 $\pm$ 0.9 & -- \\
& LLM-J & \textbf{78.4 $\pm$ 0.5} & 68.5 $\pm$ 0.4 & 62.8 $\pm$ 0.4 & 69.3 $\pm$ 0.4 & 75.8 $\pm$ 1.2 \\

\bottomrule
\end{tabular}
}
\end{table}

\section{Ablation Studies}
\label{appendix:sec:ablation:studies}
This section presents two ablation studies on the MultiModalQA dataset to isolate the contributions of EDMA's key components: entropy minimization and cross-modal conditioning.

\subsection{Ablation 1: Random Trajectory Selection}

Table \ref{tab:ablation_random_trajectory} shows the performance impact of disabling entropy minimization during trajectory selection. Instead of choosing the candidate trajectory that minimizes logical entropy, a trajectory is selected uniformly at random while keeping all other pipeline components identical. To account for the stochasticity of random selection, results are averaged over 10 runs using the LLM-Judge metric. Across all evaluated models, selecting trajectories uniformly at random causes a severe performance drop ($-32.2$ percentage points on average) compared to entropy-guided selection. This shows that EDMA's performance gains do not merely stem from task decomposition into intermediate partitions, but critically rely on selecting the trajectory that minimizes logical entropy.

\begin{table}[h]
\centering
\caption{Performance comparison between full EDMA and random trajectory selection.}
\label{tab:ablation_random_trajectory}
\resizebox{0.70\textwidth}{!}{%
\begin{tabular}{lrrr}
\toprule
\textbf{Model} & \textbf{EDMA} & \textbf{EDMA (Random Trajectory)} & \textbf{Difference} \\
\midrule
GPT-5.2            & 80.5 & 45.8 & -34.7 \\
Nova-2-Lite        & 65.9 & 47.3 & -18.6 \\
Kimi-K2.5          & 83.2 & 37.1 & -46.1 \\
Nemotron           & 75.1 & 51.6 & -23.5 \\
Qwen3              & 82.5 & 48.2 & -34.3 \\
Claude-4.6         & 88.9 & 53.0 & -35.9 \\
\midrule
\textbf{Average}   & \textbf{79.3} & \textbf{47.2} & \textbf{-32.2} \\
\bottomrule
\end{tabular}
}
\end{table}

\subsection{Ablation 2: Removing Cross-Modal Conditioning}
To isolate the contribution of cross-modal conditioning, we disable multimodal partitioning. Specifically, we execute requirement extraction, unimodal partitioning, and entropy-based trajectory selection, but omit the construction of multimodal partitions via cross-modal conditioning. Final answers are thus generated solely from the lowest-entropy unimodal partition.

\begin{table}[h]
\centering
\caption{Performance comparison between full EDMA and unimodal-only partitioning.}
\label{tab:ablation_cross_modal}
\resizebox{0.70\textwidth}{!}{%
\begin{tabular}{lrrr}
\toprule
\textbf{Model} & \textbf{EDMA} & \textbf{EDMA (Unimodal-only)} & \textbf{Difference} \\
\midrule
GPT-5.2            & 80.5 & 65.5 & -15.0 \\
Nova-2-Lite        & 65.9 & 48.0 & -17.9 \\
Kimi-K2.5          & 83.2 & 73.1 & -10.1 \\
Nemotron           & 75.1 & 56.8 & -18.3 \\
Qwen3              & 82.5 & 68.4 & -14.1 \\
Claude-4.6         & 88.9 & 77.8 & -11.1 \\
\midrule
\textbf{Average}   & \textbf{79.3} & \textbf{64.9} & \textbf{-14.4} \\
\bottomrule
\end{tabular}
}
\end{table}

As shown in Table~\ref{tab:ablation_cross_modal}, removing cross-modal conditioning leads to an average performance drop of $14.4$ percentage points across all models. This drop is of the same order as the share of questions that require cross-modal reasoning: approximately $35\%$ of questions in MultiModalQA strictly require it, whereas $\sim 65\%$ can be answered using a single modality. This is consistent with the framework failing to resolve cross-modal dependencies without cross-modal conditioning, and confirms that multimodal partitioning is fundamental for handling multimodal queries.

\section{Statistical significance}
\label{appendix:sec:statistical_significance}

\subsection{EDMA vs PP}
We establish the statistical reliability of the performance gains achieved by EDMA over PP by conducting paired statistical significance testing across all evaluated backbone architectures.

Statistical significance is assessed using both a parametric test (paired two-tailed $t$-test) and a non-parametric test (two-tailed Wilcoxon signed-rank test) at a significance threshold of $\alpha = 0.05$.

Table \ref{tab:significance} reports the statistical comparison between EDMA and PP across backbones. EDMA achieves statistically significant improvements ($p < 0.05$) over PP across every model and metric evaluated under both parametric and non-parametric tests, demonstrating robust gains across diverse model families.

\begin{table}[h]
\centering
\caption{Statistical comparison of EDMA against PP across model backbones. $\Delta$ represents the mean score difference ($\text{EDMA} - \text{PP}$). Tick (\ding{51}) indicates significance at $p < 0.05$.}
\label{tab:significance}
\resizebox{\textwidth}{!}{%
\begin{tabular}{llrrrrcc}
\toprule
\textbf{Backbone} & \textbf{Metric} & \textbf{EDMA (mean)} & \textbf{PP (mean)} & \textbf{$\Delta$} & \textbf{Sig. ($t$-test)} & \textbf{Sig. (Wilcoxon)} \\
\midrule
\multirow{3}{*}{Claude-4.6}        & EM        & 82.7 & 57.9 & $+24.9$ & \ding{51} & \ding{51} \\
                                   & F1        & 83.0 & 70.7 & $+12.4$ & \ding{51} & \ding{51} \\
                                   & LLM-judge & 88.8 & 85.4 &  $+3.3$ & \ding{51} & \ding{51} \\
\midrule
\multirow{3}{*}{Kimi-K2.5}         & EM        & 80.7 & 60.2 & $+20.5$ & \ding{51} & \ding{51} \\
                                   & F1        & 81.3 & 72.4 &  $+8.8$ & \ding{51} & \ding{51} \\
                                   & LLM-judge & 83.3 & 81.7 &  $+1.6$ & \ding{51} & \ding{51} \\
\midrule
\multirow{3}{*}{Nemotron}          & EM        & 80.0 & 51.9 & $+28.1$ & \ding{51} & \ding{51} \\
                                   & F1        & 80.7 & 61.3 & $+19.4$ & \ding{51} & \ding{51} \\
                                   & LLM-judge & 74.9 & 64.9 & $+10.0$ & \ding{51} & \ding{51} \\
\midrule
\multirow{3}{*}{Nova-2-Lite}       & EM        & 79.1 & 47.2 & $+31.9$ & \ding{51} & \ding{51} \\
                                   & F1        & 80.4 & 57.1 & $+23.3$ & \ding{51} & \ding{51} \\
                                   & LLM-judge & 66.6 & 42.1 & $+24.5$ & \ding{51} & \ding{51} \\
\bottomrule
\end{tabular}%
}
\end{table}

\subsection{Modality Fusion Decreases Entropy}
\label{appendix:sec:mod_fusion_entropy}
\begin{figure}[h]
    \centering
    \includegraphics[width=0.70\linewidth]{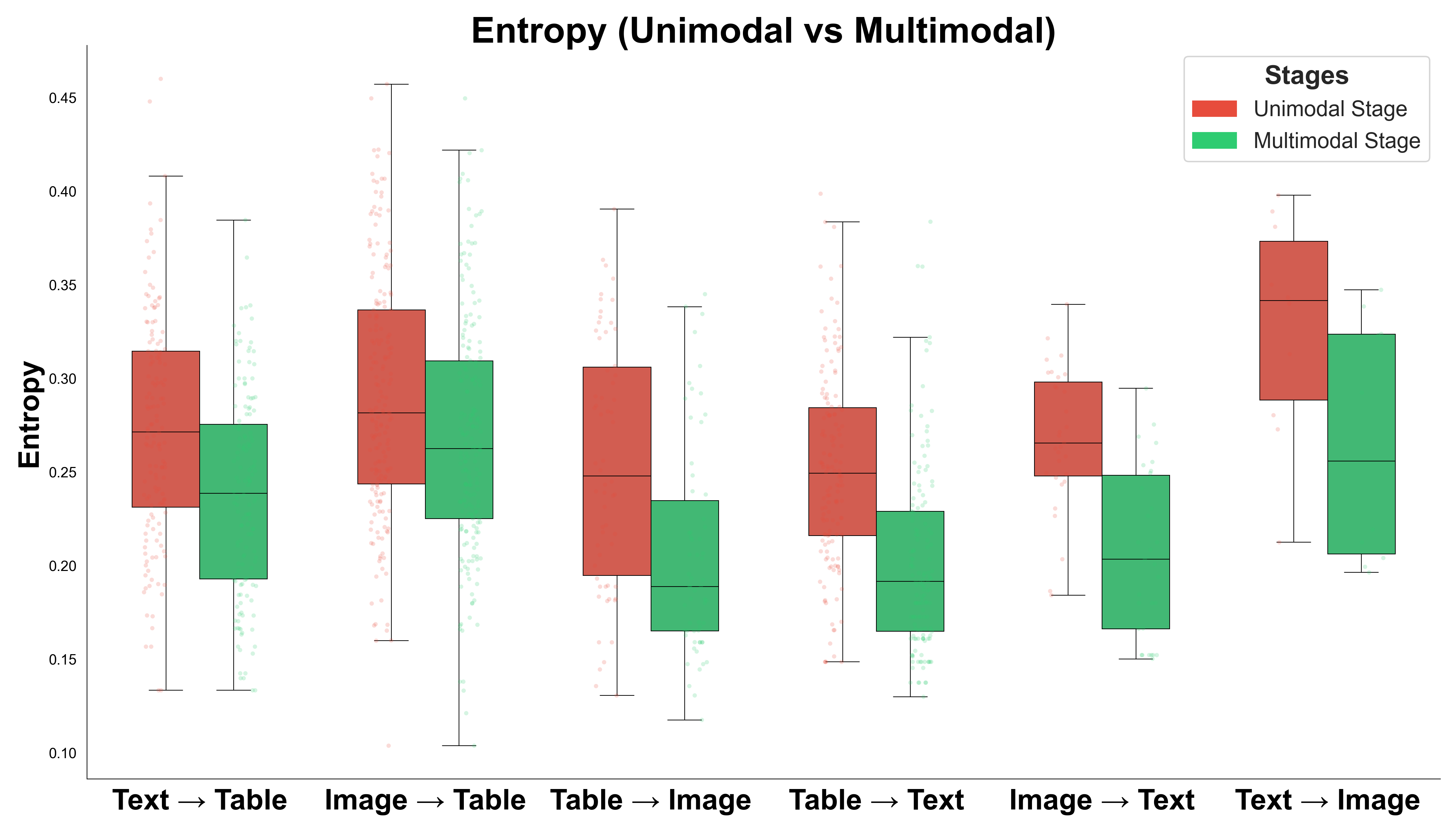}
    \caption{Multimodal entropy distribution consistently shows lower values compared to unimodal entropy distributions.}
    \label{fig:info_gain_process}
\end{figure}

We analyse the role of modality fusion in EDMA through the lens of logical entropy reduction over the candidate answer space. Figure \ref{fig:info_gain_process} compares the distribution of entropy values before and after modality fusion, which occurs when cross-modal evidence is extracted from one modality and used to condition the target modality.

Conditioning the target modality with complementary information from a secondary modality decreases logical entropy, thereby reducing uncertainty. This shows that modality fusion is associated with lower entropy, mitigating uncertainty arising from noisy or ambiguous unimodal signals. Table \ref{tab:entropy_reduction_analysis} reports the statistical significance of these reductions across all directional modality pairs.

\begin{table}[h]
\centering

\caption{Statistical Evaluation of Entropy Reduction via Information Fusion. $\overline{H}$ represents mean entropy; $t$ represents the t-statistic; Point-Biserial (PB) is the correlation between a binary modality indicator (unimodal vs. multimodal) and entropy values. Negative PB values reflect lower entropy under multimodal representations. Tick (\ding{51}) indicates significance at $p < 0.05$.}
\label{tab:entropy_reduction_analysis}
\begin{tabular}{lccccccc}
\toprule
 & & & \multicolumn{2}{c}{\textbf{T-Test}} & \multicolumn{2}{c}{\textbf{Point-Biserial}} \\
\cmidrule(lr){4-5} \cmidrule(lr){6-7}
\textbf{Combination direction} & $\boldsymbol{\overline{H(Uni)}}$ & $\boldsymbol{\overline{H(Multi)}}$ & \textbf{Stat ($t$)} & \textbf{Sig. ($p$)} & \textbf{Corr. ($r$)} & \textbf{Sig. ($p$)} \\ 
\midrule
\textbf{Image $\rightarrow$ Table} & 0.29 & 0.27 & 8.84 & \ding{51} & -0.16 & \ding{51} \\
\textbf{Image $\rightarrow$ Text}  & 0.27 & 0.21 & 5.80 & \ding{51} & -0.60 & \ding{51} \\
\textbf{Table $\rightarrow$ Image} & 0.25 & 0.21 & 6.82 & \ding{51} & -0.36 & \ding{51} \\
\textbf{Table $\rightarrow$ Text}  & 0.25 & 0.20 & 12.16 & \ding{51} & -0.43 & \ding{51} \\
\textbf{Text $\rightarrow$ Image}  & 0.33 & 0.27 & 4.11 & \ding{51} & -0.47 & \ding{51} \\
\textbf{Text $\rightarrow$ Table}  & 0.27 & 0.24 & 9.07 & \ding{51} & -0.29 & \ding{51} \\
\bottomrule
\end{tabular}
\end{table}

\subsection{Higher Entropy Is Associated with More False Positives}
\label{appendix:subsec:higher_entropy_fp}

To investigate the relationship between entropy and error rates, we compare entropy levels between correct answers (True Positives, TP) and incorrect answers (False Positives, FP). As detailed in Table \ref{tab:entropy_analysis_tp_fp}, false positives consistently exhibit higher mean entropy across all modalities.

\begin{table}[h]
\centering
\caption{Statistical comparison of entropy in True Positives (TP) vs. False Positives (FP). $\overline{H}$ denotes mean entropy; $t$ denotes the t-test statistic; PB denotes the point-biserial correlation between entropy and a binary outcome variable (TP = 0, FP = 1). Tick (\ding{51}) indicates significance at $p < 0.05$.}
\label{tab:entropy_analysis_tp_fp}
\resizebox{0.70\textwidth}{!}{%
\begin{tabular}{lccccccc}
\toprule
 & & & \multicolumn{2}{c}{\textbf{T-Test}} & \multicolumn{2}{c}{\textbf{Point-Biserial}} \\
\cmidrule(lr){4-5} \cmidrule(lr){6-7}
\textbf{Modality} & $\boldsymbol{\overline{H(TP)}}$ & $\boldsymbol{\overline{H(FP)}}$ & \textbf{Stat ($t$)} & \textbf{Sig. ($p$)} & \textbf{Corr. ($r$)} & \textbf{Sig. ($p$)} \\ 
\midrule
\textbf{Text}  & 0.20 & 0.24 & -8.72 & \ding{51} & 0.26 & \ding{51} \\
\textbf{Image} & 0.21 & 0.26 & -5.78 & \ding{51} & 0.26 & \ding{51} \\
\textbf{Table} & 0.21 & 0.27 & -8.66 & \ding{51} & 0.23 & \ding{51} \\
\bottomrule
\end{tabular}
}
\end{table}

We further assess the impact of entropy on False Discovery Rates ($\text{FDR}$) by grouping entropy into Low, Medium, and High level. Table \ref{tab:friedman_global} confirms significance via the Friedman test, while Table \ref{tab:pairwise_wilcoxon} provides pairwise Wilcoxon signed-rank comparisons across all entropy levels.

\begin{table}[h]
\centering
\caption{Statistical comparison across modalities using Mean False Discovery Rates ($\overline{\text{FDR}}$) and the Friedman Test for Related Samples. $Q$ denotes the Friedman test statistic. Tick (\ding{51}) indicates significance at $p < 0.05$.}
\label{tab:friedman_global}
\resizebox{0.70\textwidth}{!}{%
\begin{tabular}{lcccccc}
\toprule
 & \multicolumn{3}{c}{\textbf{Mean FDR}} & \multicolumn{2}{c}{\textbf{Friedman Test}} \\
\cmidrule(lr){2-4} \cmidrule(lr){5-6}
\textbf{Modality} & $\boldsymbol{\overline{\text{FDR}}_{\text{Low}}}$ & $\boldsymbol{\overline{\text{FDR}}_{\text{Medium}}}$ & $\boldsymbol{\overline{\text{FDR}}_{\text{High}}}$ & \textbf{Stat ($Q$)} & \textbf{Sig. ($p$)} \\
\midrule
\textbf{Text}  & 0.081 & 0.475 & 0.503 & 130.72 & \ding{51} \\
\textbf{Image} & 0.189 & 0.766 & 0.782 & 108.10 & \ding{51} \\
\textbf{Table} & 0.133 & 0.459 & 0.491 & 40.36  & \ding{51} \\
\bottomrule
\end{tabular}
}
\end{table}

\begin{table}[H]
\centering
\caption{Pairwise comparisons of Mean False Discovery Rates ($\overline{\text{FDR}}$) across entropy levels within each modality using the Wilcoxon Signed-Rank Test for paired samples. Tick (\ding{51}) indicates significance at $p < 0.05$.}
\label{tab:pairwise_wilcoxon}
\resizebox{0.70\textwidth}{!}{%
\begin{tabular}{lcccccc}
\toprule
 & \multicolumn{3}{c}{\textbf{Low vs Medium}} & \multicolumn{3}{c}{\textbf{Low vs High}} \\
\cmidrule(lr){2-4} \cmidrule(lr){5-7}
\textbf{Modality} 
& \textbf{Low} & \textbf{Med.} & \textbf{Sig. ($p$)}
& \textbf{Low} & \textbf{High} & \textbf{Sig. ($p$)} \\
\midrule
\textbf{Text}  
& 0.081 & 0.475 & \ding{51}
& 0.081 & 0.503 & \ding{51} \\

\textbf{Image} 
& 0.189 & 0.766 & \ding{51}
& 0.189 & 0.782 & \ding{51} \\

\textbf{Table} 
& 0.133 & 0.459 & \ding{51}
& 0.133 & 0.491 & \ding{51} \\
\bottomrule
\end{tabular}
}
\end{table}

Table \ref{tab:fdr_entropy} breaks down the False Discovery Rate ($\text{FDR}$) controlling for modality paths and candidate set size. We split candidate set sizes (ranging from 13 to 26) into three buckets containing approximately 33\% of the queries each. Even after controlling for modality path and candidate set size, $\text{FDR}$ increases monotonically moving from low to medium and high entropy.

\begin{table}[h]
\centering
\caption{False Discovery Rate across modality paths, candidate set size buckets, and entropy levels. FDR increases consistently as entropy increases from Low to Medium and High.}
\label{tab:fdr_entropy}
\resizebox{0.90\textwidth}{!}{%
\begin{tabular}{llllll}
\toprule
\textbf{Modality Pair} & \textbf{Candidate Bucket} &
\textbf{Low Entropy FDR} & \textbf{Medium Entropy FDR} &
\textbf{High Entropy FDR} \\
\midrule
Text           & 13--19 & 0.161 & 0.479 & 0.480 \\
Text           & 20--23 & 0.084 & 0.359 & 0.482 \\
Text           & 24--26 & 0.071 & 0.426 & 0.461 \\
\midrule
Text $\rightarrow$ Image & 13--19 & 0.250 & 0.933 & 1.000 \\
Text $\rightarrow$ Image & 20--23 & 0.267 & 0.917 & 1.000 \\
\midrule
Text $\rightarrow$ Table & 13--19 & 0.200 & 0.350 & 0.533 \\
Text $\rightarrow$ Table & 20--23 & 0.167 & 0.428 & 0.514 \\
Text $\rightarrow$ Table & 24--26 & 0.130 & 0.258 & 0.481 \\
\midrule
Image $\rightarrow$ Text & 13--19 & 0.200 & 0.678 & 1.000 \\
Image $\rightarrow$ Text & 20--23 & 0.167 & 0.333 & 0.433 \\
Image $\rightarrow$ Text & 24--26 & 0.173 & 0.851 & 1.000 \\
\midrule
Image          & 13--19 & 0.273 & 0.758 & 0.766 \\
Image          & 20--23 & 0.185 & 0.678 & 0.784 \\
Image          & 24--26 & 0.158 & 0.799 & 0.860 \\
\midrule
Image $\rightarrow$ Table & 13--19 & 0.175 & 0.650 & 0.823 \\
Image $\rightarrow$ Table & 20--23 & 0.272 & 0.550 & 0.821 \\
Image $\rightarrow$ Table & 24--26 & 0.217 & 0.769 & 0.879 \\
\midrule
Table $\rightarrow$ Text & 13--19 & 0.111 & 0.565 & 0.620 \\
Table $\rightarrow$ Text & 20--23 & 0.125 & 0.600 & 0.702 \\
Table $\rightarrow$ Text & 24--26 & 0.045 & 0.432 & 0.436 \\
\midrule
Table $\rightarrow$ Image & 13--19 & 0.131 & 0.889 & 0.933 \\
Table $\rightarrow$ Image & 20--23 & 0.067 & 0.675 & 0.711 \\
Table $\rightarrow$ Image & 24--26 & 0.200 & 0.583 & 0.875 \\
\midrule
Table          & 13--19 & 0.131 & 0.235 & 0.475 \\
Table          & 20--23 & 0.077 & 0.107 & 0.177 \\
Table          & 24--26 & 0.036 & 0.414 & 0.414 \\
\bottomrule
\end{tabular}%
}
\end{table}

\section{Impact Assessment}
\label{appendix:sec:impact_assessment}
The introduction of the EDMA methodology enhances the ability of LLMs to solve cross-modal problems by guiding the answering process through entropy minimization and producing a quantifiable answer trajectory. This traceability allows users to understand which modalities contributed most to the final answer, improving the interpretability and reliability of LLM predictions in real-world scenarios. By providing structured answering trajectories, EDMA also makes LLM-based systems more robust and trustworthy. While privacy and trustworthiness are general concerns associated with LLMs, our methodology does not introduce additional risks beyond those already inherent to LLMs.

\section{Prompts}
\label{appendix:sec:prompts}

\subsection{Prompts EDMA}

\subsubsection{Prompt Modality Decision}

\begin{table}[h]
\centering
\caption{Prompt used to determine which modalities are required to answer a query.}
\label{tab:modality_prompt}
\resizebox{0.89\textwidth}{!}{%
\begin{tabular}{p{0.95\linewidth}}
\toprule

\textbf{System Prompt} \\
\midrule

You are a Question Answering assistant. \\ Your task is to determine which modality or combination of modalities is required to answer a given question. You are provided with: text paragraphs, images (with titles), and tables. \\ \\ 

Select exactly one of the following labels:
\begin{itemize}
\item \texttt{image}: the answer can be derived from images alone
\item \texttt{text}: the answer can be derived from text alone
\item \texttt{table}: the answer can be derived from tables alone
\item \texttt{image\_text}: the answer requires both images and text
\item \texttt{image\_table}: the answer requires both images and tables
\item \texttt{text\_table}: the answer requires both text and tables
\item \texttt{text\_table\_image}: the answer requires text, tables, and images
\end{itemize}

Guidelines:
\begin{itemize}
\item Use only the provided information
\item Output exactly one label, with no additional text
\end{itemize}

\\
\midrule

\textbf{User Prompt} \\
\midrule

\texttt{Question: \{question\}} \\

\texttt{Images: \{images\_text\}} \\

\texttt{Text Paragraphs: \{paragraphs\_text\}} \\

\texttt{Tables: \{tables\_text\}} \\ \\ 

Select the required modality or modality combination.

\\
\bottomrule
\end{tabular}
}
\end{table}

\subsubsection{Prompt Extraction of Answer Requirements}
\begin{table}[h]
\centering
\caption{Prompt used to extract structured answer requirements from natural-language questions.}
\label{tab:criteria_prompt}
\resizebox{0.90\textwidth}{!}{%
\begin{tabular}{p{0.95\linewidth}}
\toprule

\textbf{System Prompt} \\
\midrule

You are a Question Answering assistant. \\
You task is to convert natural-language questions into structured query frames. \\ \\

Given a question, extract a structured representation with the following components:

\begin{itemize}
    \item \textbf{Topic.} Select exactly one topic from the predefined set: \texttt{\{topics\}}. Choose the most dominant topic; if unclear, use \texttt{Other}. 

    \item \textbf{Expected Answer Type.} Identify both: (i) a specific answer type, and (ii) a general entity category. Focus on the type of entity the answer represents.

    \item \textbf{Target Entity.} Identify the main entity referenced in the question.

    \item \textbf{Constraints.} Extract atomic, independently verifiable conditions from the question. Each constraint should include:
    \begin{enumerate}
    \item a constraint type (e.g., relation, award, time, role, qualifier)
    \item the exact supporting phrase from the question
    \item a normalized version of that phrase
    \end{enumerate}
\end{itemize}

\\ \midrule

\textbf{User Prompt} \\
\midrule

\texttt{Question: \{question\_text\}}

\\
\bottomrule
\end{tabular}
}
\end{table}

\subsubsection{Prompt Checking Answer Requirement Paragraph}
\begin{table}[h]
\centering
\caption{Prompt to determine the presence of content overlap between a paragraph and an answer requirement.}
\label{tab:text_filter_prompt}
\begin{tabular}{p{0.95\linewidth}}
\toprule

\textbf{System Prompt} \\
\midrule

You are a Question Answering assistant. \\ Your task is to determine whether a text paragraph (or its title) contains any concept specified in a given answer requirement. \\ \midrule

\textbf{User Prompt} \\
\midrule

\texttt{Title: \{metadata\}} \\

\texttt{Paragraph: \{text\}} \\

\texttt{Answer requirement: \{r\}} \\ \\ 

Determine whether the paragraph contains any concept from the answer requirement. If at least one concept from the answer requirement is explicitly supported by the paragraph or its title, respond \texttt{yes}; otherwise, respond \texttt{no}. \\ \bottomrule
\end{tabular}
\end{table}

\subsubsection{Prompt Checking Answer Requirement Image}
\begin{table}[h]
\centering
\caption{Prompt to determine the presence of content overlap between an image and an answer requirement.}
\label{tab:image_filter_prompt}
\begin{tabular}{p{0.95\linewidth}}
\toprule

\textbf{System Prompt} \\
\midrule

You are a Question Answering assistant. \\ Your task is to determine whether an image or its title contains any concept specified by an answer requirement.
\\ \midrule

\textbf{User Prompt} \\
\midrule

\texttt{Title: \{metadata\}} \\

\texttt{Image: \{image data\}} \\

\texttt{Answer requirement: \{r\}} \\ \\ 

If at least one concept from the answer requirement is explicitly supported by the image content or its title, respond \texttt{yes}; otherwise, respond \texttt{no}.

\\
\bottomrule
\end{tabular}
\end{table}

\subsubsection{Prompt Checking Answer Requirement Table Row}
\begin{table}[h]
\centering
\caption{Prompt to determine the presence of content overlap between a table row and an answer requirement.}
\label{tab:row_filter_prompt}
\begin{tabular}{p{0.95\linewidth}}
\toprule

\textbf{System Prompt} \\
\midrule

You are a Question Answering assistant. \\ Your task is to determine whether a table row contains any concept specified a given answer requirement. \\ \midrule

\textbf{User Prompt} \\
\midrule

\texttt{Table Row: \{table row\}} \\

\texttt{Answer requirement: \{r\}} \\ \\ 

If at least one concept from the answer requirement is explicitly supported by the row content, respond \texttt{yes}; otherwise, respond \texttt{no}. \\ \bottomrule
\end{tabular}
\end{table}

\newpage
\subsubsection{Prompt to Extract Information Text}

\begin{table}[h]
\centering
\caption{Prompt used for extracting question-relevant evidence from a paragraph.}
\label{tab:restricting_text_prompt}
\resizebox{\textwidth}{!}{%
\begin{tabular}{p{0.95\linewidth}}
\toprule

\textbf{System Prompt} \\
\midrule

You are a Question Answering assistant. \\
Your task is to extract relevant information from a given paragraph in relation to a question. You are given the question, the paragraph title, and the paragraph. Follow the instructions to extract the relevant information:

\begin{enumerate}
    \item \textbf{Step 1.} Question analysis: determine whether the paragraph is related to the question using only the provided inputs. 
    \item \textbf{Step 2.} Evidence extraction: if the paragraph is related to the question, extract the minimal information relevant to answering the question. The description must include only the entity or value that directly connects the paragraph to the question.
\end{enumerate}

\\ \midrule

\textbf{User Prompt} \\
\midrule

\texttt{Question: \{question\_text\}} \\

\texttt{Paragraph Title: \{title\}} \\

\texttt{Paragraph: \{text\}} \\ \\

Provide the relevant information according to the instructions. \\ \bottomrule
\end{tabular}
}
\end{table}

\subsubsection{Prompt to Extract Information Image}
\begin{table}[h]
\centering
\caption{Prompt used for extracting question-relevant evidence from an image.}
\label{tab:restricting_image_prompt}
\resizebox{\textwidth}{!}{%
\begin{tabular}{p{0.95\linewidth}}
\toprule

\textbf{System Prompt} \\
\midrule

You are a Question Answering assistant. \\

Your task is to extract relevant information from a given image in relation to a question. You are given the question, the image title, and the image. Follow the instructions to extract the relevant information:

\begin{itemize}
    \item \textbf{Step 1.} Question analysis: determine whether the image is related to the question using only the provided inputs.
    \item \textbf{Step 2.} Evidence extraction: if the image is related to the question, produce a short textual description of the image that captures the minimal information relevant to answering the question. The description must include only the entity or value that directly connects the image to the question. 
\end{itemize}

\\ \midrule

\textbf{User Prompt} \\
\midrule

\texttt{Question: \{question\_text\}} \\

\texttt{Image Title: \{title\}} \\

\texttt{Image: \{text\}} \\ \\

Provide the relevant information according to the instructions. \\ \bottomrule
\end{tabular}
}
\end{table}

\newpage
\subsubsection{Prompt to Extract Information Table Row}
\begin{table}[h]
\centering
\caption{Prompt used for extracting question-relevant evidence from a table row.}
\label{tab:restricting_row_prompt}
\resizebox{\textwidth}{!}{%
\begin{tabular}{p{0.95\linewidth}}
\toprule

\textbf{System Prompt} \\
\midrule

You are a Question Answering assistant. \\

Your task is to extract relevant information from a given table row in relation to a question. You are given the question and the table row. Follow the instructions to extract the relevant information:

\begin{itemize}
    \item \textbf{Step 1.} Question analysis: determine whether the table row is related to the question using only the provided inputs.
    \item \textbf{Step 2.} Evidence extraction: if the table row is related to the question, produce a short textual description of the table row that captures the minimal information relevant to answering the question. The description must include only the entity or value that directly connects the table row to the question.
\end{itemize}

\\ \midrule

\textbf{User Prompt} \\
\midrule
\texttt{Question: \{question\_text\}} \\
\texttt{Paragraph Title: \{title\}} \\
\texttt{Paragraph: \{text\}} \\ \\
Provide the relevant information according to the instructions. 
\\ \bottomrule
\end{tabular}
}
\end{table}

\subsubsection{Prompt to Check Mutual Information Between Paragraphs}
\begin{table}[h]
\centering
\caption{Prompt to determine the presence of mutual information between two paragraphs.}
\label{tab:text_bridge_prompt}
\resizebox{\textwidth}{!}{%
\begin{tabular}{p{0.95\linewidth}}
\toprule

\textbf{System Prompt} \\
\midrule

You are a Question Answering assistant. \\

You are given a question and two paragraphs. Your task is to determine whether any entity, concept, or fact mentioned in one paragraph is also mentioned in the other paragraph. \\ \midrule

\textbf{User Prompt} \\
\midrule

\texttt{Question: \{question\_text\}} \\

\texttt{Title Paragraph 1: \{title\_{1}\}} \\

\texttt{Paragraph 1: \{content\_paragraph\_1\}} \\

\texttt{Title Paragraph 2: \{title\_{2}\}} \\

\texttt{Paragraph 2: \{content\_paragraph\_2\}} \\ \\

Determine whether any entity, concept, or fact mentioned in \texttt{Paragraph 1} is also present in \texttt{Paragraph 2}. If any shared entity, name, or concept appears in both paragraphs, respond \texttt{yes}. Otherwise respond \texttt{no}. \\ \bottomrule
\end{tabular}
}
\end{table}

\newpage
\subsubsection{Prompt Check Mutual Information Between Textual Description and Image}
\begin{table}[h]
\centering
\caption{Prompt to check the presence of mutual information between an image and a paragraph.}
\label{tab:image_bridge_prompt}
\resizebox{\textwidth}{!}{%
\begin{tabular}{p{0.95\linewidth}}
\toprule

\textbf{System Prompt} \\
\midrule

You are a Question Answering assistant. \\

You are given a question, one image, and an image title and a textual description. \\

Your task is to determine whether any entity, concept, or fact mentioned in the textual description is also present in the image or image title. 

\\ \midrule

\textbf{User Prompt} \\
\midrule

\texttt{Question: \{question\_text\}} \\

\texttt{Text: \{content\}} \\

\texttt{Image Title: \{image\_title\}} \\

\texttt{Image: \{image content\}} \\ \\

Determine whether any entity, concept, or fact mentioned in the \texttt{Text} is also present in the \texttt{Image Title} or in the \texttt{Image}. If any shared entity, name, or concept appears in both the \texttt{Text} and the \texttt{Image Title} or in the \texttt{Text} and the \texttt{Image}, respond \texttt{yes}. Otherwise respond \texttt{no}.

\\ \bottomrule
\end{tabular}
}
\end{table}

\subsubsection{Prompt to Check Mutual Information Between Textual Description and Table Row}
\begin{table}[h]
\centering
\caption{Prompt to check the presence of mutual information between a table row and a paragraph.}
\label{tab:row_bridge_prompt}
\resizebox{\textwidth}{!}{%
\begin{tabular}{p{0.95\linewidth}}
\toprule

\textbf{System Prompt} \\
\midrule

You are a Question Answering assistant. \\

You are given a question, paragraph (textual description), a table description, and a table row. \\

Your task is to determine whether any entity, concept, or fact mentioned in the textual description is also present in the table row. Use the table description to understand the context. 
\\ \midrule

\textbf{User Prompt} \\
\midrule

\texttt{Question: \{question\_text\}} \\

\texttt{Paragraph: \{content\}} \\

\texttt{Table Description: \{table\_description\}} \\

\texttt{Table Row: \{table\_row\}} \\ \\

Determine whether any entity, concept, or fact mentioned in the \texttt{Paragraph} is also present in the \texttt{Table Row}. If any shared entity, name, or concept appears in both the \texttt{Paragraph} and the \texttt{Table Row}, respond \texttt{yes}. Otherwise respond \texttt{no}.

\\ \bottomrule
\end{tabular}
}
\end{table}

\newpage
\subsubsection{Check Answer in Paragraph}
\begin{table}[h]
\centering
\caption{Prompt to check whether a paragraph contains an entity that matches the answer type.}
\label{tab:check_answer_text_prompt}
\resizebox{0.85\textwidth}{!}{%
\begin{tabular}{p{0.95\linewidth}}
\toprule

\textbf{System Prompt} \\
\midrule

You are a Question Answering assistant. \\

You are given a question, a paragraph, the expected answer type. Your task is to determine whether the paragraph contains an entity that completely or partially matches the expected answer type required by the question. \\

Guidelines: 

\begin{itemize}
    \item This is not keyword matching. Perform semantic reasoning using synonyms, paraphrases, and implicit mentions

    \item Only return \texttt{TRUE} if the paragraph contains a specific entity that could fully or partially answer the question. Partial answers are acceptable if relevant.

    \item If the paragraph only provides context without an answer entity, return \texttt{FALSE}. If uncertain, return \texttt{FALSE}.

    \item If \texttt{TRUE}, extract the exact text span from the paragraph without paraphrasing or normalization. If \texttt{FALSE}, return \texttt{NONE}
\end{itemize}

\\ \midrule

\textbf{User Prompt} \\
\midrule

\texttt{Question: \{question\_text\}} \\

\texttt{Expected Answer Type: \{answer\_class\}} \\

\texttt{Paragraph: \{paragraph\_text\}} 

\\ \\

Check the presence of the answer based on the instructions provided above.

\\ \bottomrule
\end{tabular}
}
\end{table}

\subsubsection{Check Answer in Image}
\begin{table}[h]
\centering
\caption{Prompt to check if the caption or the image contains an entity that matches the answer type.}
\label{tab:check_answer_image_prompt}
\resizebox{0.85\textwidth}{!}{%
\begin{tabular}{p{0.95\linewidth}}
\toprule

\textbf{System Prompt} \\
\midrule

You are a Question Answering assistant. \\

Your task is to determine whether either the caption or the image contain an entity that matches the expected answer type required by the question. You are given the question, the image title, and an image. You must evaluate both the image title and the image. \\ \\ 

Guidelines: 

\begin{itemize}
    \item First check whether the caption explicitly contains an entity that satisfies the expected answer type. If not, check whether the image contains a visually identifiable entity that satisfies the answer type. If neither modality satisfies either answer type, return \texttt{FALSE}.
    \item If uncertain, return \texttt{FALSE}.
    \item If evidence comes from the caption, extract the exact text span. If evidence comes from the image, provide a concise visual description. If no match, the returned entity must be \texttt{NONE}.
    \item Critical instructions for the Image: Examine all parts of the image carefully, including background, corners, and small or partially obscured objects. Pay attention to subtle details such as tiny text, small objects, fine patterns, colours, positions, or partially visible entities. Do not rely only on large, obvious objects. Small details may be crucial to answer the question. Only return \texttt{TRUE} for contains if the entity is clearly present in the image. If uncertain, return \texttt{FALSE}.
\end{itemize} 

\\ \midrule

\textbf{User Prompt} \\
\midrule

\texttt{Question: \{question\_text\}} \\

\texttt{Expected Answer Type: \{answer\_class\_specific\}} \\

\texttt{Image Title: \{caption\_text\}} \\ \\

Check the presence of the answer in the image and its title based on the instructions provided above.
\\ \bottomrule
\end{tabular}
}
\end{table}

\newpage
\subsubsection{Check Answer in Table Row}
\begin{table}[h]
\centering
\caption{Prompt to verify whether a table row contains a value that directly answers a question.}
\label{tab:check_answer_row_prompt}
\resizebox{0.90\textwidth}{!}{%
\begin{tabular}{p{0.95\linewidth}}
\toprule

\textbf{System Prompt} \\
\midrule

You are a Question Answering assistant. \\

Your task is to determine whether a table row contains a value that directly answers a question. \\

You are given the question, the expected answer type, the table description, and one table row. \\ \\

Guidelines: 

\begin{itemize}
    \item Use the description of the table to interpret the row correctly
    \item Do not use only literal meaning. Also use synonyms and contextual meaning
    \item Only return \texttt{TRUE} if one of the row's values directly answers the question
    \item If the row contains related information but no direct answer, return \texttt{FALSE}
    \item If uncertain, return \texttt{FALSE}
    \item If \texttt{TRUE}, extract the exact cell value from the row that contains the answer. If \texttt{FALSE}, return \texttt{NONE}
\end{itemize}

\\ \midrule

\textbf{User Prompt} \\
\midrule

\texttt{Question: \{question\_text\}} \\
\texttt{Expected Answer Type: \{answer\_class\_specific\}} \\
\texttt{Table Description: \{table\_description\}} \\
\texttt{Table Row: \{row\}} \\ \\

Check the presence of the answer in table row based on the instructions provided above.

\\ \bottomrule
\end{tabular}
}
\end{table}

\subsection{Baselines Prompts}

\subsubsection{Prompt for Direct Prompting}

\begin{table}[h]
\centering
\caption{Direct prompting baseline used in our experiments.}
\label{tab:dp_prompt}
\resizebox{0.90\textwidth}{!}{%
\begin{tabular}{p{0.95\linewidth}}
\toprule

\textbf{System Prompt} \\
\midrule

You are a multimodal question answering assistant. \\ Your task is to answer a given question using three types of provided data sources: text paragraphs, images (with titles), and tables. Carefully review all sources and derive the answer by combining relevant information across modalities. \\ \\

The final response should be concise and contain only the answer, without any explanation. If the necessary information is not available, output \texttt{None}. \\

\midrule

\textbf{User Prompt} \\
\midrule

\texttt{Question: \{question\_text\}} \\

\texttt{Images: \{images\_text\}} \\

\texttt{Text Paragraphs: \{paragraphs\_text\}} \\

\texttt{Tables: \{tables\_text\}} \\ \\

Provide the final answer to the question. \\ 
\bottomrule
\end{tabular}
}
\end{table}

\newpage
\subsubsection{Prompt for Chain-of-Thought}
\begin{table}[h]
\centering
\caption{Chain-of-Thought (CoT) prompt used in our experiments.}
\label{tab:cot_prompt_readable}
\resizebox{0.78\textwidth}{!}{%
\begin{tabular}{p{0.95\linewidth}}
\toprule

\textbf{System Prompt} \\
\midrule

You are a multimodal question answering assistant. Your task is to answer a given question using only the provided information sources:
text paragraphs, images (with titles), and tables. You must not rely on any external or prior knowledge. If the provided information is insufficient, output \texttt{None}. \\ \\ 

Procedure to follow:

\begin{itemize}
    \item \textbf{Step 1.} Extract information: Identify relevant facts from each modality: text (key statements), images (objects, labels, relationships), and tables (rows, columns, values).

    \item \textbf{Step 2.} Align information: Match entities and concepts that appear across different modalities and explicitly link them.

    \item \textbf{Step 3.} Integrate information: Combine the aligned evidence to form a complete understanding, ensuring that all parts of the question are addressed. 
    
    \item \textbf{Step 4.} Derive the answer: Formulate the final answer, clearly grounded in the combined evidence from the modalities. 
    \item \textbf{Step 5.} Verify: Ensure that the answer is fully supported by the provided information and that no relevant modality has been ignored. 
\end{itemize}

Output the reasoning separated from the final response. The final response should be concise and contain only the answer, without any explanation.

\\ \midrule

\textbf{User Prompt} \\
\midrule

\texttt{Question: \{question\_text\}} \\

\texttt{Images: \{images\_text\}} \\

\texttt{Text Paragraphs: \{paragraphs\_text\}} \\

\texttt{Tables: \{tables\_text\}} \\ \\

Answer the question following the procedure above. \\
\bottomrule
\end{tabular}
}
\end{table}

\subsubsection{Prompt for Planning Prompting}

\begin{table}[h]
\centering
\caption{Planning-based prompting baseline used in our experiments.}
\label{tab:pp_prompt}
\resizebox{0.78\textwidth}{!}{%
\begin{tabular}{p{0.95\linewidth}}
\toprule

\textbf{System Prompt} \\
\midrule

You are a multimodal question answering assistant. Your task is to answer a given question using the provided data sources: text paragraphs, images (with titles), and tables. You must rely only on this information and not use any external knowledge. If the provided information is insufficient, output \texttt{None}. \\ \\ 

Before producing the final answer, follow the three phases below:

\textbf{Phase 1: Strategic Planning} \\
Analyse the question and available data sources, and devise a step-by-step plan. The plan should:
\begin{itemize}
\item Identify which data sources are most relevant
\item Determine the order in which they should be processed
\item Specify how to handle potential conflicts across sources
\end{itemize}

\textbf{Phase 2: Execution} \\
Follow the plan to extract and combine information. Every piece of evidence must be explicitly grounded in a source (e.g., table, image title, or paragraph). \\

\textbf{Phase 3: Final Answer} \\
Provide a concise answer based on the executed plan. If the required information is not available, output \texttt{None}. \\
\midrule

\textbf{User Prompt} \\
\midrule

\texttt{Question: \{question\_text\}} \\

\texttt{Images: \{images\_text\}} \\

\texttt{Text Paragraphs: \{paragraphs\_text\}} \\

\texttt{Tables: \{tables\_text\}} \\ \\

Follow the planning and execution procedure to answer the question. \\
\bottomrule
\end{tabular}
}
\end{table}



\end{document}